\documentclass{article}
\usepackage{caption}
\usepackage{microtype}
\usepackage{graphicx}
\usepackage{subfigure}
\usepackage{booktabs} 
\usepackage{enumitem}

\usepackage{hyperref}
\usepackage{silence}
\newcommand{\sib}{SiBBlInGS}

\usepackage{algorithm}

\usepackage[accepted]{icml2024}

\usepackage{amsmath}
\usepackage{amssymb}
\usepackage{mathtools}
\usepackage{amsthm}
\usepackage{bm}

\usepackage[capitalize,noabbrev]{cleveref}

\theoremstyle{plain}

\theoremstyle{definition}

\theoremstyle{remark}

\usepackage[textsize=tiny]{todonotes}

\icmltitlerunning{SiBBlInGS: Similarity-driven Building-Block Inference using Graphs across States}

\begin{document}

\twocolumn[
\icmltitle{SiBBlInGS: Similarity-driven Building-Block Inference using Graphs across States}



\begin{icmlauthorlist}
\icmlauthor{Noga Mudrik}{yyy}
\icmlauthor{Gal Mishne}{comp}
\icmlauthor{Adam S. Charles}{yyy}
\end{icmlauthorlist}
]

\icmlaffiliation{yyy}{Biomedical Engineering, Kavli NDI, Center for Imaging Science, The Mathematical Institute for Data Science, The Johns Hopkins University,
      Baltimore, MD, USA}
\icmlaffiliation{comp}{Halıcıoğlu Data Science Institute,
UCSD, 
San Diego, CA, USA}

\icmlcorrespondingauthor{Noga Mudrik}{nmudrik1@jhu.edu}
\icmlcorrespondingauthor{Gal Mishne}{gmishne@ucsd.edu}
\icmlcorrespondingauthor{Adam Charles}{adamsc@jhu.edu}

\icmlkeywords{dictionary learning, multi-way data;  neural variability, graph-driven learning; regularization; sparse representation}
\vskip 0.3in



\printAffiliationsAndNotice{}  
\begin{abstract}
Time series data across scientific domains are often collected under distinct states (e.g., tasks), wherein latent processes (e.g., biological factors) create complex inter- and intra-state variability. A key approach to capture this complexity is to uncover fundamental interpretable units within the data, Building Blocks (BBs), which modulate their activity and adjust their structure across observations. Existing methods for identifying BBs in multi-way data often overlook inter- vs. intra-state variability, produce uninterpretable components, or do not align with properties of real-world data, such as missing samples and sessions of different duration. Here, we present a framework for Similarity-driven Building Block Inference using Graphs across States (SiBBlInGS). SiBBlInGS offers a graph-based dictionary learning approach for discovering sparse BBs along with their temporal traces, based on co-activity patterns and inter- vs. intra-state relationships. Moreover, SiBBlInGS captures per-trial temporal variability and controlled cross-state structural BB adaptations, identifies state-specific vs. state-invariant components, and accommodates variability in the number and duration of observed sessions across states. We demonstrate SiBBlInGS's ability to reveal insights into complex phenomena as well as its robustness to noise and missing samples through several synthetic and real-world examples, including web search and neural data.
\end{abstract}

\section{Introduction}

The analysis of high-dimensional time-series is increasingly important across various scientific disciplines, ranging from neuroscience~\cite{kala2009fuzzy,mudrik2022decomposed} to social sciences~\cite{jerzak2023improved} to environmental studies~\cite{hipel1994time}. 
These data, however, present a daunting challenge in terms of comprehensibility as they are often highly heterogeneous. 
Specifically, data in many domains are gathered under multiple states (e.g., clinical interventions), while latent factors may introduce variability across trials within states (e.g., internal biological processes that lead to variations in patient responses to treatment).
Current analysis methods often struggle to capture the full variability in such multi-state data. Additionally, 
integrating data from repeated observations (trials) under the same state into a coherent representation is often challenged by missing samples or variable trial duration and sampling rates~\cite{goris2014partitioning,charles2018dethroning, duncker2018temporal}. The common practice of within-state trial averaging, for example, obscures important patterns within individual trials. 

\begin{figure*}[ht!]
\centering
\includegraphics[width=1\textwidth]{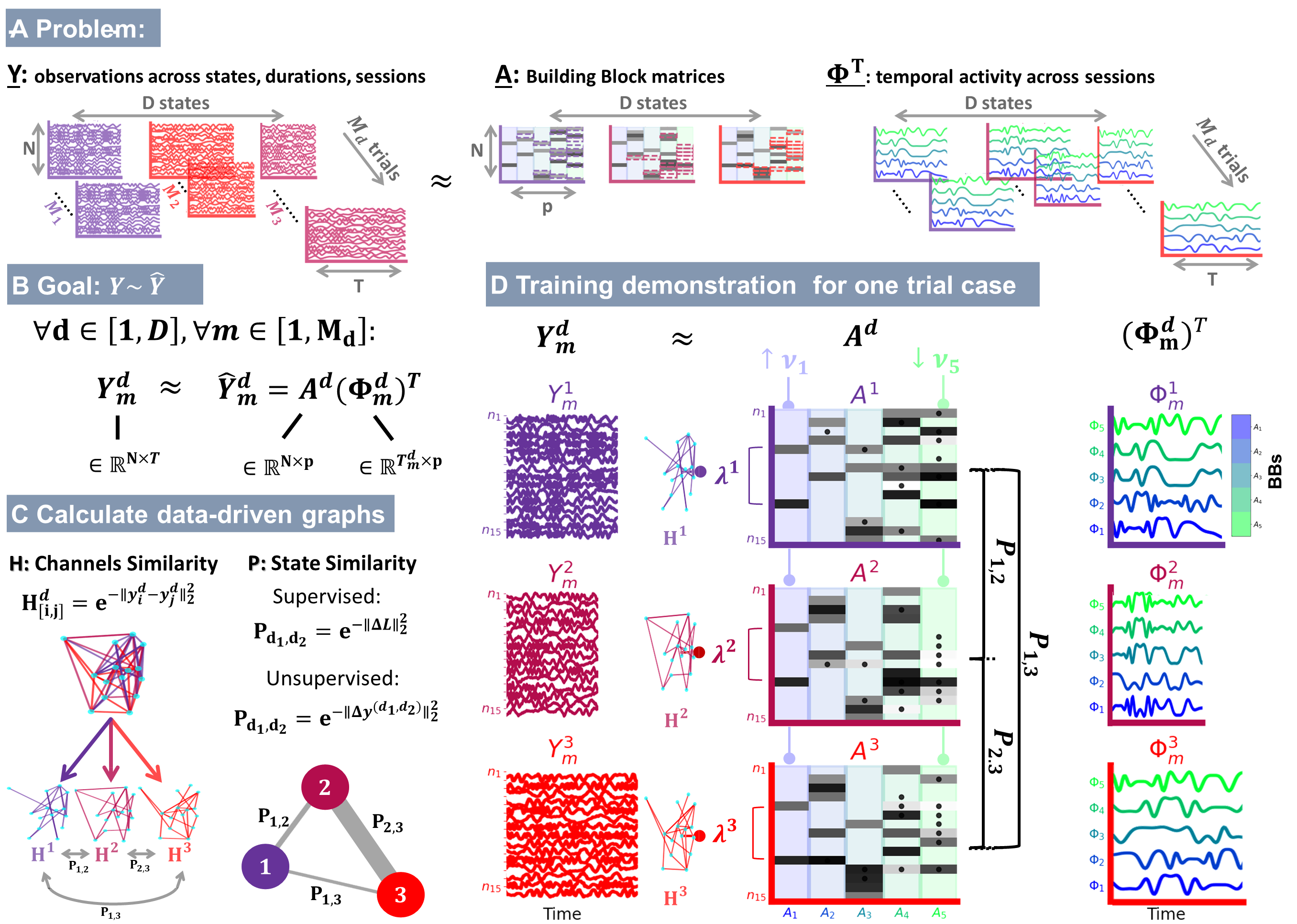}
\caption{\textbf{SiBBlInGS Schematic} 
\textbf{A} 
SiBBlInGS adapts to real-world datasets with varying session durations, sampling rates, and state-specific data by learning interpretable graph-driven hidden patterns and their temporal activity.
\textbf{B} SiBBlInGS is based on a per-state-and-trial matrix factorization where the BBs ($\bm{A}^d$) are identical across trials and similar across states.
\textbf{C} SiBBlInGS controls the BB similarity via data-driven channel graphs ($\bm{H}^d \in \mathbb{R}^{N \times N }$) and a state similarity graph ($\bm{P} \in \mathbb{R}^{D \times D}$), which can be either predefined (supervised) or data-driven. 
\textbf{D} The learning schematic with an exemplary trial for each of the 3 exemplary states. 
The BBs of each state $d$ (columns of $\textbf{A}^d$) are
constrained with two regularization terms: 1) state-specific $\bm{\lambda}^d$ captures similar activity between channels by leveraging the channel-similarity graph $\bm{H}^d$, and 2) $\bm{P}$, captures BB consistency across states via the state similarity graph.
$\bm{\nu}$ controls the relative level of cross-state similarity between BBs, allowing the discovery of both background and state-specific BBs. 
Higher (lower) $\bm{\nu}$ values promote greater (lesser) consistency of specific BBs across states (e.g. $\bm{\nu}_1$ v.s $\bm{\nu}_5$). }
\label{fig:intro_fig}
\end{figure*}

A promising approach for analyzing multi-state data involves identifying fundamental representational units---Building Blocks (BBs)---whose composition remains similar across states, while their temporal profiles can modulate across trials to capture trial-to-trial variability, both within the same state and across states. 
These BBs can represent, for instance, neural ensembles in the brain; social groups in diverse contexts; gene clusters under regulatory mechanisms, etc.
Identifying these BBs and understanding how they change across states is a key step for recognizing the latent processes underlying the data and providing valuable insights into core commonalities and differences among states. 
However, uncovering these BBs poses a challenge, as their individual activities or compositions are often unobservable. 
This challenge is further complicated by potential variations across states, not only in the temporal activity of these BBs, but also in the subtle structural adjustments of the BBs' compositions between states. 
For example, a neural ensemble may display varying temporal activity between normal (non-seizure) brain activity sessions and during seizures, along with subtle structural adaptations in the ensemble composition during seizures~\cite{vandenberg2008probing}, e.g., neurons that are not typically part of the ensemble might become involved during a seizure.



Here, we present SiBBlInGS, a graph-based data driven framework to unravel the complexities of high-dimensional multi-state time-series data, by unveiling its underlying sparse, similarity-driven BBs along with their temporal activity. Our main contributions include:
\begin{itemize}[noitemsep]
\item We develop a novel framework to find interpretable hidden BBs underlying high-dimensional multi-way data while extracting their 
cross-trial temporal activities
and inter-state structural variability, and address real-world challenges unmet by existing methods. 
\item We accommodate varying trial conditions, including different time durations, sampling rates, missing samples, and per-state counts, and enable overlapping BB composition.
\item We highlight our method's promise by demonstrating its ability to recover ground-truth components in synthetic data and meaningful latents in several real-world examples. 
\end{itemize}

\section{Background and Related Work}

In the case of single-trial analysis,
methods for identifying BBs often rely on matrix decomposition including Singular Value Decomposition (SVD,~\citet{kogbetliantz1955solution}), Principal Components Analysis (PCA,~\citet{hotelling1933analysis}), Independent Components Analysis (ICA,~\citet{hyvarinen2001independent}), or Non-negative Matrix Factorization (NMF,~\citet{lee1999learning}), where sparsity constraints can be added to improve interpretability, e.g., sparse PCA (SPCA,~\citet{zou2006sparse}). Extending these to the multi-trial setting can be addressed by either concatenating trials end-to-end to create a single wide matrix or by applying these methods individually to each trial. However, this either overlooks the temporal scales of the data (within trial and cross-trial) or ignores shared factors across trials. 

A more suitable extension to multi-trial observations is tensor factorization (TF) methods, e.g., PARAFAC and Tucker decomposition~\cite{harshman1970foundations,Williams2018,mishne2016, de2000multilinear,wu2018neural}, which consider the trials as an additional dimension of the data. However, none of these methods nor their combinations with Gaussian processes (GP,~\citet{tillinghast2020probabilistic, xu2011infinite, zhe2016distributed}) and dynamic information (e.g., NNDTN and NONFAT,~\citet{wang2022nonparametric}) can naturally handle variability in trial duration or address state variability as a fourth dimension.

Extensions of TF methods seek either identical BBs across states with flexible temporal patterns or fixed temporal traces across states with flexible cross-state BBs. For example, the Shared Response Model (SRM)~\cite{NIPS2015_b3967a0e} models similar temporal activity across individuals in multi-subject fMRI studies while accounting for varying spatial responses between subjects. SRM, however, requires that components be orthogonal, which may not align with biological plausibility. Hyperalignment (HA)~\cite{haxby2011common} addresses a similar setting as SRM by rotating the subjects’ time series responses to optimize inter-subject correlation. However, HA does not explicitly reduce the dimension of the feature space.

Other existing methods, such as Dynamic Mode Decomposition (DMD)~\cite{schmid2010dynamic}, model the temporal dynamics explicitly as dynamical systems, however, these methods are tailored for 2D analysis and thus are not designed to simultaneously model data that vary both within and between states. State-Space Models (SSMs)~\cite{auger2021guide} represent another approach to explore time-series data by describing the latent states' evolution by a state-transition matrix; however, they do not aim to find sparse interpretable ensembles with cross-trial structural and temporal variability. 
Other methods include demixed PCA (dPCA)~\cite{brendel2011demixed}, Targeted Dimensionality Reduction (TDR)~\cite{mante2013context} and model-based TDR (mTDR)~\cite{aoi2018model,aoi2020prefrontal}. The latter two directly regress rank-1 (TDR) or low-rank (mTDR) components that explicitly target task-relevant variables. However, TDR/mTDR similarly cannot handle trials of varying duration and do not incorporate sparsity in the identified ensembles. dPCA falls short in addressing missing data, different trial durations, and varied sampling rates.

Fuzzy clustering~\cite{yang1993survey,9072636,HE2018109} allows data points to exhibit varying degrees of membership in multiple clusters, addressing limitations of methods that restrict data points to a single BB. 
However, these approaches focus solely on BB structures rather than their temporal activities, and do not integrate within-state and between-state variability information.

Closer to our approach, dictionary learning~\cite{olshausen2004sparse,olshausen1996wavelet,aharon2006k},
provides more interpretable representations~\cite{tovsic2011dictionary}
by learning a feature dictionary  where each data point can be linearly reconstructed using only a few of the feature vectors. 
While traditional dictionary learning treats each data point as independent, recent advances based on re-weighted $\ell_1$~\cite{candes2008enhancing,garrigues2010group} can account for spatio-temporal similarities in the sparse feature representations between data points~\cite{garrigues2010group,charles2013spectral, charles2016dynamic, zhang2011sparse, qin2017stochastic,mishne2019learning}. 
In particular, re-Weighted $\ell_1$ Graph Filtering (RWL1-GF)~\cite{charles2022graft} was recently developed for demixing fluorescing components in calcium imaging recordings by correlating the sparse decompositions across a data-driven graph defined by pixel similarity.
While RWL1-GF proves the efficacy of graph embeddings in extracting meaningful features, it is constrained to single-trial data and confines its graph construction to a single dimension of the data---the pixel space---overlooking possibly  meaningful structures in other dimensions.

More advanced methods, including nonlinear deep learning models, have also been developed to extract latent factors. However, these typically lack interpretability in mapping back to the feature/sample space, are limited in their ability to produce sparse latent factors, and require large amounts of training data. 
Variational Autoencoders (VAEs)~\cite{xu2018spherical,tillinghast2021nonparametric} offer recovery of nonlinear latent low-dimensional representations; however, they do not naturally consider the data's temporal structure, and their elements do not directly represent the contribution of individual channels from the input space. 
Sparse~\cite{ashman2020sparse,barello2018sparse} or dynamical~\cite{girin2020dynamical} variants of VAEs do not consider within-vs between state variability of the latent representations.
Transformer models, e.g.,~\cite{liu2022seeing} jointly model individual and collective dynamics via an individual module for each of several component dynamics and an interaction module that captures pairwise interactions. However, this necessitates prior knowledge about the system's separation and requires large-scale data due to the encoder-decoder architecture.
Recently, for neuronal data analysis, CEBRA~\cite{schneider2023learnable} incorporates auxiliary labels and temporal information in contrastive optimization, however, it produces a latent state that requires additional interpretation steps to connect to the neuronal space.

%

Notably, all the above approaches are constrained in their capacity to identify fundamental hidden sparse components while capturing multi-state, multi-trial variability.



\section{Problem Definition and Notations}

Consider a system with $N$ channels organized into at most $p$ BBs, with each BB representing a group of channels with shared functionality. These BBs serve as the fundamental constituents of a complex process, however their composition is not directly observed nor explicitly known.
In particular, let the columns of $\bm{A} \in \mathbb{R}^{N \times p}$ represent the BBs, such that $\bm{A}_{ij}$ is the contribution of the $i$-th channel to the $j$-th BB, with $\bm{A}_{ij} = 0$ indicating that channel $i$ does not belong to BB $j$. We further assume that each channel can sparsely belong to multiple BBs with varying degrees of membership, such that $\|\bm{A}_{i:}\|_0 = K < p \textrm{ for all } i = 1\dots N$. 

First, we consider a single trial of the system $\bm{Y} \in \mathbb{R}^{N \times T}$ over $T$ time points. During this trial, each BB exhibits temporal activity denoted by $\bm{\Phi} \in \mathbb{R}^{T \times p}$, that might reflect current hidden properties of the system, where $\bm{\Phi}_{t,j}$ is the activity of the $j$-th BB at time $t$. 
These temporal profiles are assumed to be smooth, bounded (i.e., $||\bm{\Phi}||_F < \epsilon_1$, for some $\epsilon_1$), and have a low correlation between distinct BBs' activity (i.e., $\rho(\bm{\Phi}_{:j}, \bm{\Phi}_{:i}) < \epsilon_2 \quad \forall{i \neq j}$, for some $\epsilon_2$). 
In this single trial case, our observations, \(\bm{Y}\), arise from the collective activity of all BBs operating together, \({\bm{Y} = \bm{A}{\bm{\Phi}}^T + \eta}\), where \(\eta\) denotes \textit{i.i.d.} Gaussian observation noise.
However, the individual composition (\(\bm{A}\)) or activity (\(\bm{\Phi}\)) of each BB is unknown.

In the more general setting, we observe a set of $M$ trials, $\{\bm{Y}_m\}_{m=1}^{M}$, where the duration of each trial $m = 1\dots M$ may vary, i.e., $\bm{Y}_m \in \mathbb{R}^{N \times T_m}$. The BBs ($\bm{A}$) remain constant across trials while their corresponding temporal activity ($\{\bm{\Phi}_m\}_{m=1}^M \textrm{ s.t. } \bm{\Phi}_m \in \mathbb{R}^{T_m \times p}$) may vary across trials to capture trial-to-trial variability.

The setting we focus on extends beyond a single set of trials; instead, we deal with a collection of $D$ such multi-trial sets, where each set is associated with a known state $d = 1\dots D$~(Fig.~\ref{fig:intro_fig}A). Across these sets, both the number of trials per set ($M_d$ for each set $d$), and the durations of the trials, may vary.
Thus, the full observation dataset includes the collection of $D$ multi-trial sets, 
$\{\bm{Y}^1_{m}\}_{m = 1}^{M_1},...\{\bm{Y}^D_{m}\}_{m = 1}^{M_D}$, each representing a different state $d = 1 \dots D$, such that ${\bm{Y}_{m}^d=\bm{A}^{d} ({\bm{\Phi}_m^d})^T + \eta_m^d}$.

We assume that the BBs' temporal activities ($\{\bm{\Phi}_m^d\}$) can vary between trials, both within and between states, and the compositions of the BBs ($\{\bm{A}^d\}$) might present subtle controlled adaptations between, but not within, states. 

Specifically, we posit that the BBs' dissimilarity between any pair of distinct states $d$ and $d'$ reflects the dissimilarity between those states, such that the distance between $\bm{A}^d$ and $\bm{A}^{d'}$  is constrained by $\|\bm{A}^d - \bm{A}^{d'}\|_F < \epsilon_3(d,d')$ for some threshold $\epsilon_3(d,d')$ determined by the application. 
For example, if considering different disease stages as states, we assume that consecutive disease stages are more similar to each other than to a healthy state, such that $\epsilon_3(d_{\textrm{stage}_{\textrm{1}}}, d_{\textrm{stage}_\textrm{2}}) < \epsilon_3 (d_{\textrm{healthy}}, d_{\textrm{stage}_{\textrm{1}}})$.
Another example is neural development, where dynamic processes drive gradual changes in neuron connectivity~\cite{golshani2009internally, marom2002development}, potentially yielding ensembles whose dissimilarity  ($\|\bm{A}^d - \bm{A}^{d{\text{\textquoteright}}}\|_F^2$)  in stages $d \neq d{\text{\textquoteright}}$ is constrained by the stages' distance 
$\epsilon_3(d, d{\text{\textquoteright}})$.

The main challenge that SiBBlInGS addresses is recovering the unknown BBs ($\bm{A}^d$) and their temporal activities ($\bm{\Phi}_m^d$) for all states and trials given only their combined simultaneous activity (Fig.~\ref{fig:intro_fig}A,B).
\begin{figure*}[t!]
\centering
\includegraphics[width=1\textwidth]{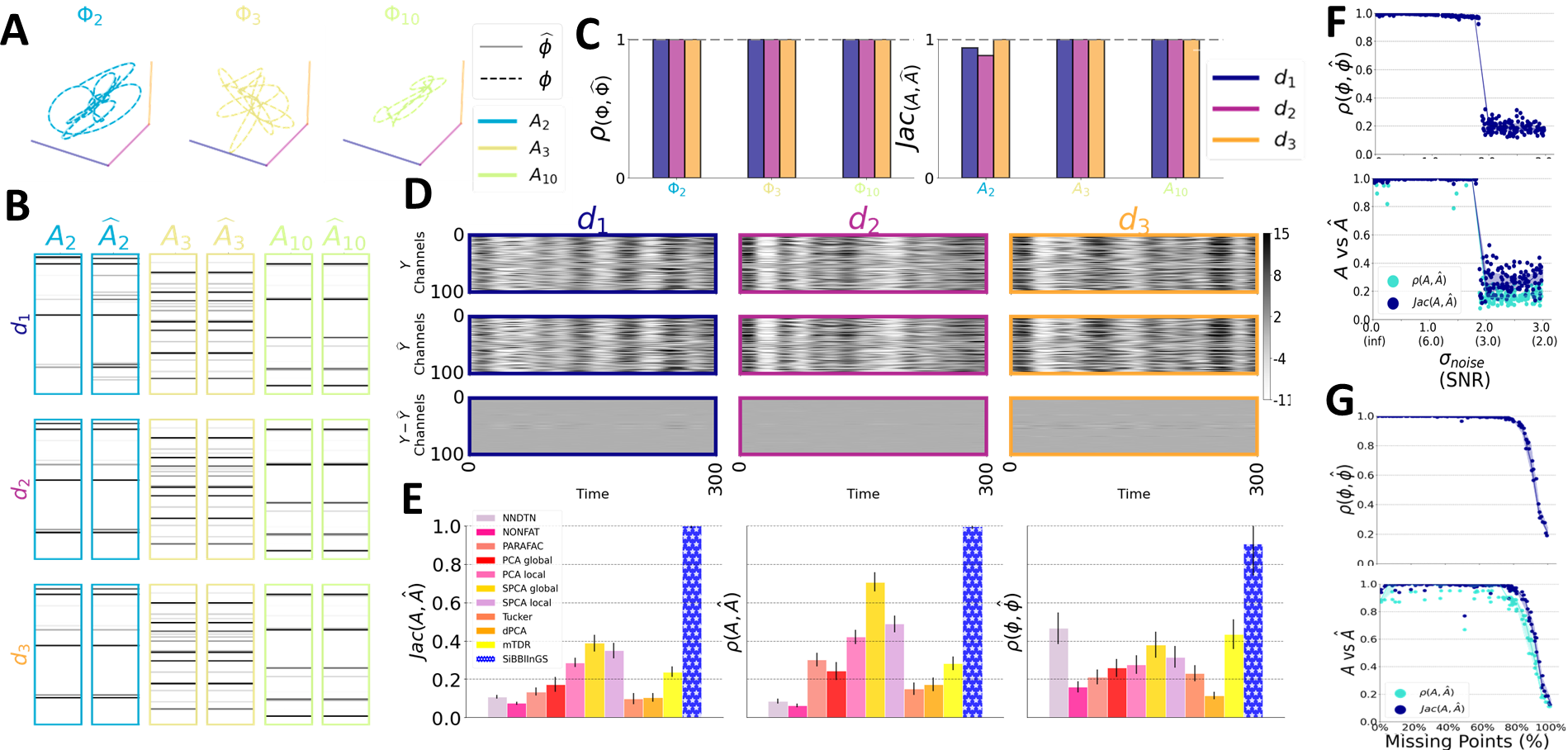}
\caption{\textbf{Synthetic data results.}
\textbf{A} Three example time traces identified by SiBBlInGS vs. ground truth traces, projected into the three synthetic states.  SiBBlInGS recovers both traces that are highly correlated with specific states (e.g., $\bm{\Phi}_{10}$; green), as well as traces that exhibit similar activation across states (e.g., $\bm{\Phi}_{2}$; blue).
\textbf{B} Comparison between the identified example BBs and the ground-truth BBs.
\textbf{C} Correlation between the example identified time traces and the ground truth (left), and Jaccard index of the identified BBs compared to the ground truth (right).
\textbf{D} Comparison between the ground-truth data (top), SiBBlInGS reconstruction  (middle), and the residual data (bottom).
\textbf{E} Comparison to baseline methods (Sec.~\ref{sec:exp}, App.~\ref{sec:comp_parafac}).
\textbf{F} Performance under noise and random initializations (300 repetitions). Each dot is a model instance. The curve shows the median values, and the shading corresponds to the 25\%-75\% percentiles. While SiBBlInGS remains robust under varying noise ($\sigma_{\textrm{signal}} / \sigma_{\textrm{noise}} >3$), it experiences a phase transition at a specific noise level, aligning with the dictionary-learning literature (e.g.~\cite{studer2012dictionary}).
\textbf{G} Performance with increasing levels of missing samples (200 repeats). The scattered dots represent model repetitions, the curves depict the median values calculated by rounding to the nearest 5\%, and the background shading corresponds to 25\%-75\% percentiles.
}
\label{fig:synth}
\end{figure*}

\section{SiBBlInGS}
In this section, we present SiBBlInGS---our framework to identify interpretable BBs along with their temporal traces based on shared activation patterns across trials and states.
Unlike existing methods, SiBBlInGS identifies BBs in high-dimensional data based on temporal similarity without assuming orthogonality, 
enables BB interdependency or overlap, and can tackle trials of different duration, sampling rates, or trial count per state (Tab.~\ref{tab:comp_table}).
SiBBlInGS provides the flexibility to select either a data-driven unsupervised approach or leverage expert-based knowledge (if desired) for integrating inter-state similarities, thus offering tailored solutions based on the specific needs of the data.


SiBBlInGS is based on an extended dictionary learning-like procedure that alternates between updating the BBs ($\{\bm{A}^d\}$) and their temporal profiles ($\{\bm{\Phi}^d_{m}\}$) for all states. 
Critical to our approach is the integration of the non-linear similarities between both channels and states.
We capture these relationships via two graphs (Fig.~\ref{fig:intro_fig}C), one over channels ($\bm{H}^d \in \mathbb{R}^{N\times N}$) to identify cross-channel regularities, and one over states ($\bm{P} \in \mathbb{R}^{D \times D}$) to promote cross-state similarity in BB structure. Mathematically, we formulate the fit $\{\widehat{\bm{A}}^d\}, \{ \widehat{\bm{\Phi}}^d_m\}$ for all $d = 1 \dots D$ and $m = 1 \dots M_d$
by minimizing the cost function
\begin{align}
  \min_{\substack{\{\bm{A}^d\}, \{\bm{\Phi}_m^d\}}}  &\sum_{d}^D \left( \sum_{m}^{M_d} \left[ \|\bm{Y}_m^d - \bm{A}^d (\bm{\Phi}_m^d)^T\|_F^2  + \mathcal{R}(\bm{\Phi}^d_m)  \right] + \right. \nonumber \\
& \left.   \mathcal{R}(\bm{A}^d)   +  \sum_{d'\neq d}^D P_{d,d'}\|(\bm{A}^d - \bm{A}^{d'}) \bm{V}\|_F^2 \right) \nonumber 
\end{align}
where the first term is a data fidelity term  and the second term regularizes the BBs' temporal traces. The term  $\mathcal{R}(\bm{A}^d)$ regularizes each BB to be a sparse group of channels based on shared temporal patterns, and the last term regularizes the BBs' similarity across states (Fig.~\ref{fig:intro_fig}).
The use of $\bm{V} = \textrm{diag}({\bm{\nu}}) \in \mathbb{R}^{p \times p}$, accompanied by the weight vector ${\bm{\nu}} \in \mathbb{R}^{p}$, allows assigning varying weights to cross-state BBs' similarities to facilitate the discovery of state-invariant vs. state-specific BBs.

SiBBlInGS thus iteratively updates \(\bm{A}^d\) and \(\bm{\Phi}_m^d\) for each trial and state, as detailed below (a concise summary of the method is presented in Alg.~\ref{alg:algorithm_SiBBlInGS} and illustrated in Fig.~\ref{fig:intro_fig}, with the computational complexity discussed in App.~\ref{sec:model_complexity}):



\noindent\textbf{Updating $\bm{A}^d$ :} 
\\
Since we assume that BBs may require subtle state-to-state adaptations but remain constant within a state,
SiBBlInGS demands that the BB matrix ($\bm{A}^d$) is shared between trials of the same state but can undergo subtle adjustments between states, proportionate to the similarities of the corresponding states as captured by $\bm{P}$.
The update of ${\bm{A}^d}$ for each state $d$, is achieved via an extended re-weighted $\ell_1$ graph filtering with an integration of the channel-similarity graph ($\bm{H}^d ~\in~\mathbb{R}^{N \times N}$) in a way that promotes the grouping of channels with similar temporal activity onto the same BBs. 
In each updating iteration of $\bm{A}^d$, 
as a pre-calculation step, we first horizontally concatenate the observations from all \(M_d\) trials of that \(d\) state to receive the matrix ${\bm{Y}^{d{\ast}}} \in \mathbb{R}^{N \times (\sum_{m = 1}^{M_d} T_m^d)}$, and vertically concatenate the last estimates of the temporal traces from all trials of that state to build the matrix \(\bm{\Phi}^{d\ast} \in \mathbb{R}^{(\sum_{m = 1}^{M_d} T_m^d) \times p}\).

We then update each row $n = 1\dots N$ of $\widehat{\bm{A}}^d$ $(\widehat{\bm{A}}_{{n :}}^d)$ 
via a re-weighted procedure that alternates between updating $\widehat{\bm{A}}_{nj}^d$ and $\bm{\lambda}^d_{n,j}$:
\begin{align}
          \widehat{{\bm{A}}}_{n:}^d=&\arg \min_{\bm{A}_{n:}^d }  \|\bm{Y}^{d\ast}_n - \bm{A}_{n:}^d (\bm{\Phi}^{d\ast})^T\|_2^2  \label{eqn:AUpdate1}  
            + \\
& \sum_{j =1}^p \bm{\lambda}^d_{n,j}  |\bm{A}_{n,j}^d|    + \sum_{d' \neq d} \bm{P}_{dd'}\|( \bm{A}_{n:}^d - \bm{A}_{n:}^{d'} ) \circ \bm{\nu} \|_2^2\nonumber,           
\end{align}
and
\begin{align}
 \bm{\lambda}^d_{n,j} = \frac{\epsilon}{\beta + |\widehat{\bm{A}}_{n,j}^d| +  w_\textrm{graph} | \bm{H}^d_{n :}  \widehat{\bm{A}}_{: j}^d |}.\label{eqn:updateLambda}   
\end{align}

Above,  $\circ$ is element-wise multiplication, and $\beta$, $\epsilon$, and $w_\textrm{graph}$ are model hyper-parameters. The matrices $\bm{H}^d$ and $\bm{P}$ are channel and state similarity graphs, described below:

\textbf{State similarity graph:} 
\\
$\bm{P} \in \mathbb{R}^{D \times D}$ is a state similarity graph that determines the effect of the similarity between each pair of states on the regularization on the distance between their BB representations.

The default option SiBBlInGS offers for calculating $\bm{P}$ is a data-driven approach (referred to as ``unsupervised $\bm{P}$'') that automatically captures relationships within the data without requiring domain knowledge. Alternatively, if users possess knowledge about labels associated with each state, they can opt for a ``supervised $\bm{P}$'' to leverage this domain knowledge or address advanced questions regarding a sequence of states. For instance, if the states represent various neuro-developmental stages, each associated with a numerical label, and users are interested in leveraging this sequence knowledge to discover neural adjustments throughout development, they may prefer the supervised approach. 

Each of these two options (data-driven vs. supervised) offers unique benefits: 
the  unsupervised variant leverages the data itself to learn similarities and patterns without preconceived biases, whereas the
supervised variant enables explicit regulation of similarity and the incorporation of human-expert knowledge into the model.

\begin{figure*}[t!]
\centering
\includegraphics[width=1\textwidth]
{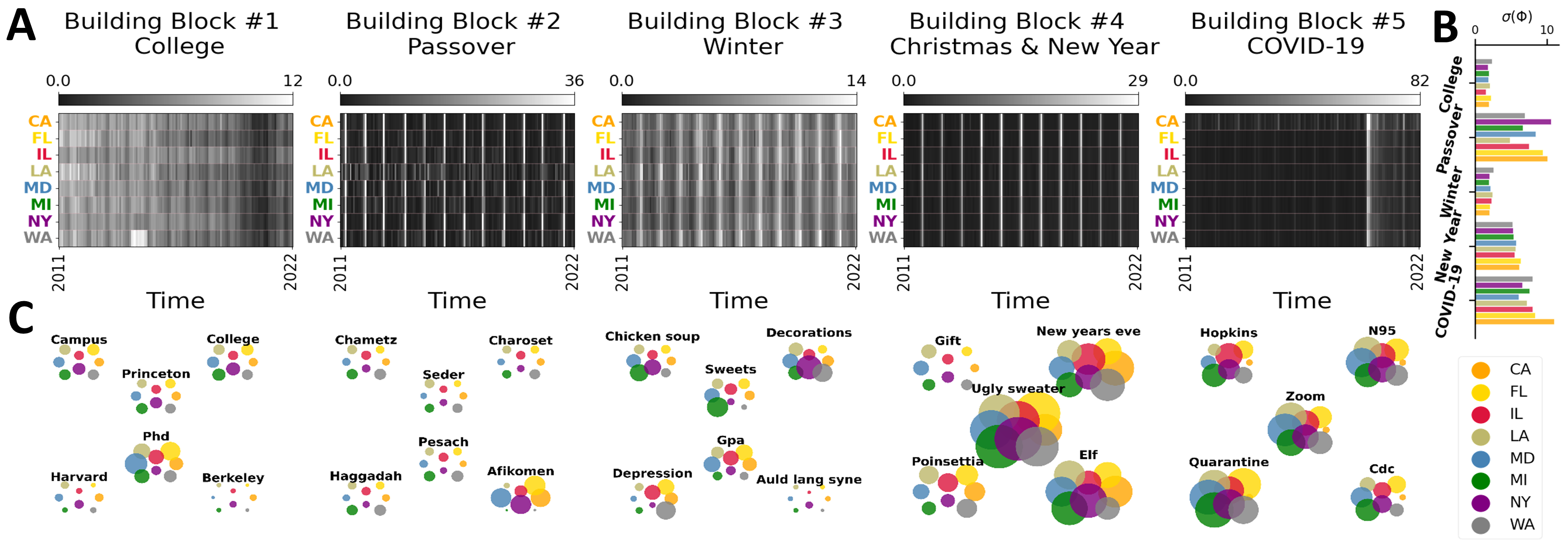}
\caption{\textbf{Demonstration on Google Trends Data.} 
\textbf{A} The BBs' temporal traces, as SiBBlInGS found, demonstrate seasonal trends consistent with the terms associated with each BB.
\textbf{B} Standard deviation of temporal traces over time for the different states align with  variability in the states' demographics (Sec.~\ref{sec:exp}).  
\textbf{C} The BBs SiBBlInGS identified along with their per-state dominancy produce more meaningful clusters than baselines (Fig.~\ref{fig:trends_comparison}). States are marked by colors; dot sizes represent the contribution of a term in the BB.  
}
\vspace{-0.2cm}
\label{fig:TRENDS}
\end{figure*}

Here, we present the supervised version of \(\bm{P}\), while the data-driven approach is detailed in Appendix B.2. This supervised version, unlike the data-driven option, assumes that each state ${d = 1, \ldots, D}$ has a numerical label \(\bm{L}^d\) (either scalar or vector) that provides valuable information for constructing the state similarity graph \(\bm{P}\).
For instance, these labels can be vectors that denote x-y coordinates in a movement task or scalars representing the speed of an object under different conditions.
 In this way, the similarity $\bm{P}_{d,d'}$ between each pair of states ($d$, $d'$) is calculated based on the distance between the labels ($\bm{L}_d$, $\bm{L}_{d'}$) associated with these states:
$ \bm{P}_{d,d'} = \exp{ \left({-\Vert\bm{L}_{d} - \bm{L}_{d'}\Vert_2^2}/{\sigma_{\bm{P}}^2} 
  \right) } $
where $\sigma_P^2$ controls how the similarities in labels scale to similarities in BBs. 
This supervised approach easily extends to both data with identical or different session duration and can also handle categorical states as described in the Appendix~\ref{sec:supervised_equal_case}. 

\textbf{Channel similarity graphs:}
\\
$\bm{H}^d \in \mathbb{R}^{N\times N}$ is the channel graph for each state $d$ and is calculated by 
${\bm{H}_{i,j}^d = \exp\left(-{\|\bm{Y}^{d\ast}_{i :} - \bm{Y}^{d\ast}_{j :}\|_2^2}/{\sigma_{\bm{H}}^2}\right) }$, 
where 
$\sigma_{\bm{H}}$ is an hyperparameter that controls the kernel bandwidth and $\bm{Y}^{d\ast}_{j :}$ is the horizontally concatenated observations under state $d$ described before.
To enhance the robustness of $\{\bm{H}^d\}_{d=1}^D$, we add a post-processing step and utilize the state-graph ($\bm{P}$) to re-weigh each $\bm{H}^d$ along the states dimension:
 ${\bm{H}^d \leftarrow \frac{\sum_{{d'} = 1}^{D} \bm{P}_{d,d'} \bm{H}^{d'}}{\|\bm{P}_d\|_1}}$. We then retain only the $k$ largest values in each row while setting the rest to zero, symmetrize, and row normalize $\bm{H}^d$ so that each row sums to one (App.~\ref{sec:channel_H}).
This process mitigates the influence of outliers and encourages the clustering of similarly-behaving channels into the same BB. 

The advantage of graph-driven re-weighting, compared to other TF and dictionary learning procedures, is that the updated weighted regularization ($\bm{\lambda}^d \in \mathbb{R}^{N \times p}$) promotes the grouping (separating) of channels with similar (dissimilar) activity into the same (different) BBs by integrating the channel similarity graph $\bm{H}^d$ into the regularization. Specifically, in the last term of the $\bm{\lambda}^d_{n,j}$'s denominator, for a given state $d$, a strong (weak) correlation between the temporal neighbors of the $n$-th channel (captured by $\bm{H}_{n :}^d$) and the members of the $j$-th BB ($\widehat{\bm{A}}_{: j}^d$) results in a decreased (increased) $\bm{\lambda}^d_{n,j}$. Consequently, the $\ell_1$ regularization on $\widehat{{\bm{A}}}_{n:}^d$ is reduced (increased)---promoting the inclusion (exclusion) of each channel into BBs that include (exclude) its temporal neighbors.

After each update of all rows in $\bm{A}^d$, each column is normalized to have a maximum absolute value of $1$.
In practice, we update $\bm{A}$ (Eq. ~\eqref{eqn:AUpdate1}) for a random subset of trials in each iteration to improve robustness and computational speed.

\noindent\textbf{Updating $\bm{\Phi}^d_m$:} 
\label{update_over_phi}
\\
The update step over $\bm{\Phi}^d_m$ uses the current estimate of $\bm{A}^d$ to re-estimate the temporal profile matrix $\bm{\Phi}^d_m$ independently over each state $d$ and trial $m$. Note that we do not enforce cross-trial similarity in $\bm{\Phi}^d_m$ to allow for flexibility in capturing trial-to-trial variability both within and across states. Thus, for each trial $m$ and state $d$, $\bm{\phi} = \bm{\Phi}_m^d$ is updated by solving:
\begin{align}
    \widehat{\bm{\phi}} =& \arg\min_{\bm{\phi}\geq 0}  \left\|\bm{Y}_m^d - \bm{A}^d {\bm{\phi}}^T \right\|_F^2 + \gamma_1\|\bm{\phi}\|_F^2  +\label{eqn:update_phi}  \\ &\gamma_2    \|\bm{\phi} - \widehat{\bm{\phi}}^{\textrm{iter}-1}\|_F^2 + 
        \gamma_3 \|\bm{\phi} - {\bm{\phi}}^{t-1}\|_F^2 
       + \gamma_4 \mathcal{R}_{\textrm{corr}}(\bm{\phi}) \nonumber
\end{align}
where the first term preserves data fidelity, 
the second term regularizes excessive activity, the third term encourages continuity across iterations 
($\widehat{\bm{\phi}}^{\textrm{iter}-1}$ is $\bm{\phi}$ from the previous iteration), and the fourth term is a diffusion term that promotes temporal consistency of the dictionary ($\bm{\phi}^{t-1}$ is $\bm{\phi}$ shifted by one time point).  ${\mathcal{R}_{\textrm{corr}}(\bm{\phi}) = \|\left( \bm{\phi}^T\bm{\phi} -\operatorname{diag}(\bm{\phi}^T\bm{\phi})\right) \circ \bm{D} \|_\textrm{sav}}$
promotes decorrelation of distinct temporal traces, where ``$\textrm{sav}$'' is sum-of-absolute-values and ${\bm{D} \in \mathbb{R}^{p \times p}}$ is a normalization matrix with ${\bm{D}_{ij} = \frac{1}{\| \bm{\phi}_{:i} \|_2 \| \bm{\phi}_{:j} \|_2}}$ (App.~\ref{solve_phi}). 
\begin{algorithm}[h!]
\caption{The SiBBlInGS Model (concise  version)}\label{alg:algorithm_SiBBlInGS}
\begin{algorithmic}
\STATE \textbf{Inputs}
\\ $\{\bm{Y}_m^{1}\}_{m=1}^{M_1},...\{\bm{Y}_m^D\}_{m=1}^{M_D}$ \COMMENT{Observations} 
\\ $\beta,\epsilon,\gamma_1,\gamma_2,\gamma_3,\gamma_4,w_{graph},\sigma_p,\sigma_H$ \COMMENT{Hyperparameters}

\STATE \textbf{Initialization and pre-Calculations}
\\$\{{\bm{A}^d\}, {\{\bm{\Phi}^d_m\}},  \forall d=1 \dots D}$ \COMMENT{Initialize BBs \& traces}

\STATE $\bm{P} \in \mathbb{R}^{D \times D}$  \COMMENT{Calculate state-similarity graph}
\STATE$\{\bm{H}^d\}_{d=1}^D$   \COMMENT{Calculate channels graphs} 
\REPEAT
    \FOR{$d = 1\dots D$} 

        \STATE Select a random batch of trials from state $d$ 

        \STATE Update $\bm{A}^d$ and $\bm{\lambda}^d$ 
         \COMMENT{via Eq.~\eqref{eqn:AUpdate1} and Eq.~\eqref{eqn:updateLambda}} 

    \FOR{$m = 1 \dots M_d$} 
            \STATE Update $\bm{\Phi}_m^d$ \COMMENT{via Eq.~\eqref{eqn:update_phi}}
    \ENDFOR
    \ENDFOR

\UNTIL{BBs and traces of all states converged}
\end{algorithmic}
\end{algorithm}

\section{Experiments}\label{sec:exp}

\subsection*{SiBBlInGS recovers ground truth BBs in synthetic data:}
Synthetic data were generated with $D = 3$ states, each consisting of a single trial, with $p = 10$ ground-truth BBs, and $N = 100$ channels. Each $i$-th BB was generated with a maximum cardinality of $\max_{d,i}{\| A_{:,i}^d \|_0} = 21$ active channels, and on average each channel was associated with $2.1$ BBs.
While the BBs were designed to be non-orthogonal, we constrained their pairwise correlations to be below a threshold of $0.6$ (${\rho}(\bm{A}_{:i}, \bm{A}_{:j}) <= 0.6$). The temporal dynamics of the synthetic data were generated by summing 15 trigonometric functions with different frequencies (App.~\ref{sec:synth_gen} for details).
\\
SiBBlInGS demonstrated monotonically improving performance during training (Fig.~\ref{fig:MSE_time}A-D), and at convergence was able to successfully recover the underlying BBs in the synthetic data 
and their temporal traces (Fig~\ref{fig:synth}{A-C}). 
In particular, example traces demonstrate a high precision of the recovered temporal traces, with correlation to the ground truth traces being close to one (Fig.~\ref{fig:synth}{A, C}, ~\ref{fig:MSE_time}F).
Furthermore, the identified BB components
align closely with the ground-truth BBs (Fig.~\ref{fig:synth}{B,C}), as indicated by high Jaccard index values. 
We compared SiBBlInGS to existing methods, 
including Tucker Decomposition, PARAFAC, (S)PCA ``global'' (a single (S)PCA for all states), (S)PCA ``local'' ((S)PCA for each state), 
NONFAT, NNDTN, mTDR, and dPCA, 
with details in App.~\ref{sec:comp_parafac}).
Notably, SiBBlInGS outperforms existing baselines both in terms of identifying the ground truth BBs and their traces (Fig.~\ref{fig:synth}{E}, ~\ref{fig:MSE_time}F,~\ref{fig:BBs_themselves},~\ref{icml-historical}).

\begin{figure*}[t!]
\centering
\includegraphics[width=1\textwidth]{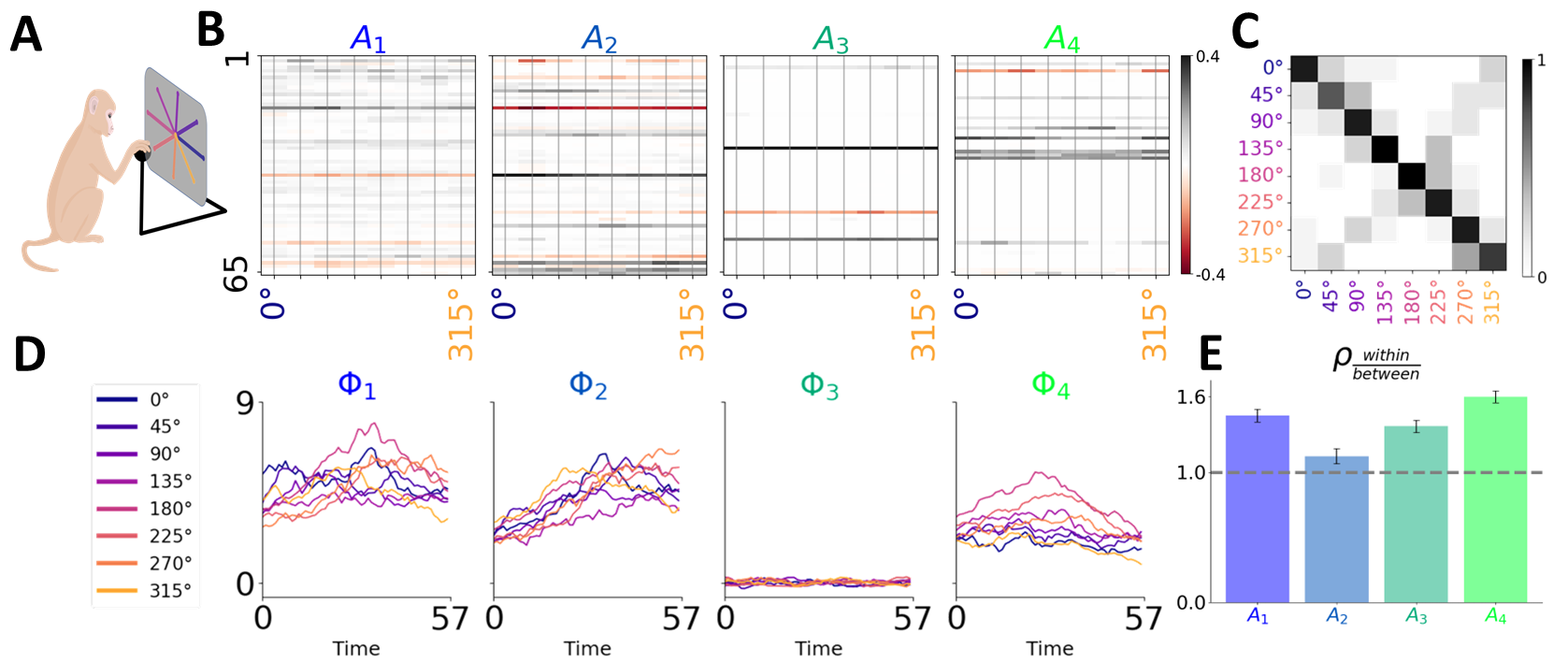}
\caption{\textbf{Identification of Temporal Patterns in Monkey Somatosensory Cortex.} 
\textbf{A} The reaching out task (\cite{andrea_colins_rodriguez_2023}). 
\textbf{B} Sparse clusters of neurons representing the identified BBs. 
\textbf{C} Confusion matrix of a multi-class logistic regression model using the inferred temporal traces to predict the state label. 
 \textbf{D} The BBs' temporal traces as they vary across states and time.
 \textbf{E}
 Ratios of within-to-between states temporal correlations for each BB, with $\frac{\rho_{\textrm{within}}}{\rho_{\textrm{between}}} > 1$, indicating states distinguishability.}
\label{fig:neuro_fig2}
\end{figure*}

\begin{figure*}[th]
\centering
\includegraphics[width=1\textwidth]{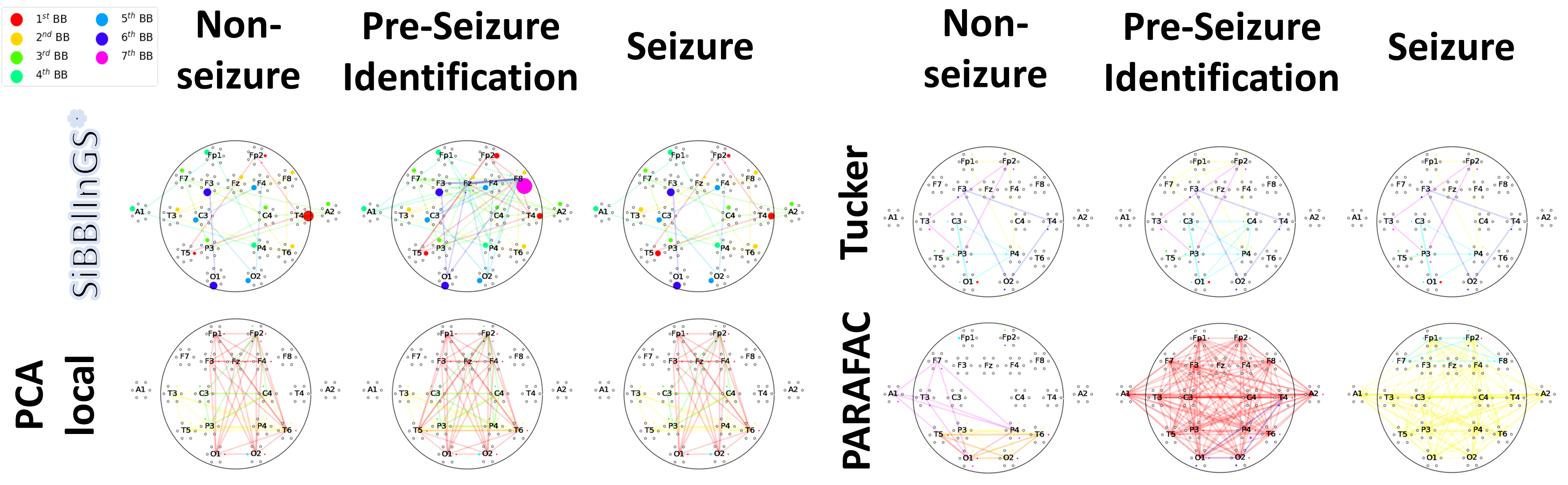}
\caption{\textbf{Emerging local BBs in Epilepsy}. The recovered BBs under 1) normal activity, 2) activity during the 8 seconds proceedings CPS seizures located around the F8 area, and 3) activity during the seizures. Colors represent different BBs, and the size of the dots corresponds to the contribution of the respective electrode to each BB. 
}
\label{fig:epi}
\end{figure*}

\subsection*{SiBBlInGS finds interpretable BBs in Google Trends:} 
We used Google Trends to demonstrate SiBBlInGS' capability in identifying temporal and structural patterns by querying search term frequency on Google over time.
We used a normalized monthly Trends volume of 44 queries (from Jan. 2011 to Oct. 2022) related to various topics, as searched in 8 US states selected for their diverse characteristics~\cite{coulby2000beyond} (see pre-processing in App.~\ref{sec:trends_pre}). 
The $p=5$ BBs identified by SiBBlInGS reveal meaningful clusters of terms
(Fig.~\ref{fig:TRENDS}B,~Fig.~\ref{fig:trends_table}), whose time traces convey the temporal evolution of user interests per region (Fig.~\ref{fig:TRENDS}A), 
while aligning with the seasonality of the BBs' components. For instance, the first BB represents college-related terms and shows a gradual annual decrease with periodic activity and a notable deviation during the COVID pandemic, possibly reflecting factors such as the shift to remote learning (Fig.~\ref{fig:TRENDS}A,~\ref{fig:trends_college}).
The second and third BBs, respectively, demonstrate periodic patterns associated with Passover in April (Fig.~\ref{fig:trends_passover}) and winter terms in December. 
 Interestingly, CA, FL, MD, and NY---all states with larger Jewish populations~\cite{pew_research_religion}---show more pronounced peaks of the ``Passover" BB activity in April (when Passover is celebrated) compared to the other states (Fig.~\ref{fig:TRENDS}B and ~\ref{fig:trends_passover}).
The last BB represents COVID-related terms and exhibits temporal patterns with a sharp increase around Jan. 2020, coinciding with the onset of the COVID pandemic in the US. 
Remarkably, while 'Hopkins' exhibits strong membership to the COVID BB in most states, it shows a less pronounced COVID-related search peak in MD (blue), where the university and hospital are located. This is likely attributed to its well-established local presence in MD (Fig.~\ref{fig:TRENDS}C, right), which contributes to the general familiarity with Hopkins throughout the year. In other states, however, Hopkins became more famous during COVID, leading to a significant nation-wide surge in Hopkins-related searches at the onset of the outbreak and contributing to its strong membership to the COVID BB in these states.
This emphasizes our model's interpretability and the need to capture similar yet distinct BBs across states.
Other methods applied with the same number of BBs as used in SiBBlInGS ($p =5$)  produced less meaningful BBs (Fig.~\ref{fig:trends_comparison}).

\subsection*{SiBBlInGS identifies meaningful patterns in brain recordings:}
We tested SiBBlInGS on neural activity (by~\citet{chowdhury2022dandiarchive}) recorded in the somatosensory cortex of a monkey performing a reaching task. 
The data we used include $M_d=18$ trials under each of the $D=8$ hand directions, with each direction corresponding to a unique state~(Fig.~\ref{fig:neuro_fig2}A).
The raw spike data were convolved over time with a Gaussian kernel to obtain firing rate estimation. 
When applying SiBBlInGS with a maximum of $p=4$ BBs, it identified sparse functional BBs (Fig.~\ref{fig:neuro_fig2}B) 
along with meaningful temporal traces (Fig.~\ref{fig:neuro_fig2}D) that exhibit state-specific patterns. 
Interestingly, the third BB consistently shows minimal activity across all states, suggesting that it captures background or noise activity. The structure of the identified BBs 
exhibits subtle yet significant adaptations across states in terms of neuron weights and BB assignments. Furthermore, SiBBlInGS finds neurons belonging to multiple neural clusters, suggesting their involvement in multiple functions.
When examining the temporal correlations of the corresponding BBs within and between states, all BBs exhibited a within/between correlations ratio  $>1$ (Fig.~\ref{fig:neuro_fig2}E,~\ref{fig:neuro_supp_fig}C, App.~\ref{sec:within_bet}) indicating robust within state trajectories and distinctions between states. Furthermore, multi-class logistic regression based only on the identified temporal traces accurately predicted the states (Fig.~\ref{fig:neuro_fig2}C).

\subsection*{SiBBlInGS discovers emerging BBs preceding seizure:}
We applied SiBBlInGS to analyze EEG recordings by~\citet{Nasreddine2021EpilepticED} from an 8-year-old epileptic patient who experienced five complex partial seizures (CPS) localized around electrode F8, as detailed in App.~\ref{sec:epi_compare}. 
SiBBlInGS unveiled interpretable and localized EEG activity in the period preceding seizures (Fig.~\ref{fig:epi}), a feat not achieved by other methods. 
It identified a BB specific to the region around the clinically labeled area (F8) that emerged during the 8 seconds prior to the seizure (Fig.~\ref{fig:epi}, pink circle in SiBBlInGS's middle).
Additionally, it found several alterations in the BB composition during the seizure in comparison to the normal activity.
E.g., the contribution of T4 to the red BB during normal activity is higher than its contribution during a seizure, while the contribution of T5 to the same BB is larger during a seizure.
This underscores the potential of SiBBlInGS in discovering BBs that uniquely emerge under specific states, made possible by the flexibility of $\bm{\nu}$ to support both state-variant and state-invariant BBs.

\section{Conclusion}

We propose SiBBlInGS for graphs-driven identification of interpretable cross-state BBs with their temporal profiles in multi-way time-series data---providing insights into systems' structure and variability. 
Unlike other approaches, SiBBlInGS naturally supports the discovery of BBs with subtle changes in cross-state structures, allows each channel to belong to a few BBs with varying contributions, and promotes the discovery of both state-invariant and state-specific BBs, while accommodating real-world variations in trial structures (e.g., varying trial duration both within and between states).
We demonstrate SiBBlInGS's capacity to identify functional neural ensembles and discern cross-state variations in web-search data structures, showcasing its promise to additional domains. 
Particularly, SiBBlInGS can be applied to any time-series data collected under various states, potentially (but not limited to) repeated measurements within each state, where the subgroups underlying the observations can undergo cross-state adjustments.
This could encompass a wide range of applications beyond those demonstrated in the paper.

\textbf{Limitations and future work:}

SiBBlInGS assumes Gaussian statistics, yet Poisson statistics may sometimes be more suitable. 
Additionally, exploring advanced distance metrics ~\cite{mishne2017data,lin2023hyperbolic} for the construction of the graphs holds promise for future research. 

The current framework assumes linearity in how the joint BBs' activity translates to the observations. While this linearity assumption is common in scientific modeling (e.g., \cite{aoi2018model, NIPS2015_b3967a0e}) due to its simplicity, versatility, direct interpretability, and biological plausibility (e.g.,~\citet{liu2019accurate}), we believe that extending the model to a nonlinear one (e.g.,~\citet{mishne2019co}) is an enticing direction for future work. 
Another direction for future extensions includes expanding the framework to accommodate missing channels or cross-trial disparities in channel identity (e.g., neural data that involve observations from different neurons across  recording sessions,~\citet{mudrik2024creimbo}).
Finally, SiBBlInGS is currently not intended to identify directional connectivity either between or within BBs  (e.g., via switched~\cite{linderman2016recurrent} or decomposed~\cite{mudrik2022decomposed, mudrik2024linocs, Yezerets2024.05.31.596903} dynamical systems prior over the temporal traces)---presenting an exciting future direction.

\section*{Acknowledgments}
N.M. was supported by the Kavli Discovery Award of the Kavli Foundation.
G.M. was partially supported by NIH grant R01EB026936.

\section*{Impact Statement}
This paper presents work that aims to advance the field of Machine Learning and Time Series Data Analysis. There are several potential societal consequences of our work, none of which we feel must be specifically highlighted here.
\bibliography{main}
\bibliographystyle{icml2024}

\newpage
\appendix
\onecolumn
\section*{Appendix}
\section{Notations}\label{sec:notation}
Throughout the paper, we adopt the following notations: the superscript $()^d$ refers to state $d$, and the subscript $()_m$ refers to trial $m$. Specifically, $\bm{Y}_m^d$ and $\Phi_m^d$ denote the observations and temporal traces of trial $m$ of state $d$, while $\bm{A}^d$ represents the BBs of state $d$.
Additionally, for a general matrix $\bm{Z}$, $\bm{Z}_{i :}$ ($\bm{Z}_{ : j}$) denotes its $i$-th row ($j$-th column). 

\begin{table}[h!]
\centering
 \hspace*{-0.5cm} 
    \caption{Notations used in the paper.}
    \begin{tabular}
    {|p{0.25\linewidth}|p{0.6\linewidth}|}     
        \hline        
        \textbf{Symbol} & \textbf{Description} \\
        \hline
        BBs &  Building Blocks \\ 
        channels & Each feature in the observations, e.g., neurons in recordings \\
        states & Different ``views" of the observations. e.g., different cognitive tasks \\
        trials/sessions & Repeated observations within state  \\
        $p$ & Number of BBs \\ 
        $D$ & Number of states \\
        $M_d$ & Number of trials for state $d$ \\ 
        $N$ & Number of channels \\
        $\bm{Z}_{n:}$ (or $\bm{Z}_{[n,:]}$) & The $n$-th row of a general matrix $\bm{Z}$ \\
        $\bm{Z}_{:i}$ (or $\bm{Z}_{[:,i]}$) & The $i$-th column of a general matrix $\bm{Z}$ \\
        $L_d$ & Label of state $d$ (optional, can be a scalar or a vector) \\
        $\bm{Y}_m^d \in \mathbb{R}^{N \times T_m^d}$ & Observation for trial $m$ and state $d$ \\            
        $\bm{A}^d \in \mathbb{R}^{N \times p}$ & Matrix of BBs for state $d$. \\ 
        $\Phi_m^d \in \mathbb{R}^{ T_m^d \times p}$ & Matrix of temporal traces for trial $m$ of state $d$.  \\
        $\bm{P} \in \mathbb{R}^{D \times D}$ & States similarity graph \\
        $\{\bm{H}^d \}_{d=1}^D$, s.t. $\bm{H}^d \in \mathbb{R}^{N \times N}$ & Channel similarity graphs \\        
        $\bm{\nu} \in \mathbb{R}^{p}$ & Controls the relative level of cross-state similarity for each BB \\
        $\bm{V} = diag(\bm{\nu})$ & A diagonal matrix whose entry in index $ii$ is the $i$-th entry of $\bm{\nu}$ \\
        $\epsilon$, $\beta$, $w_\textrm{graph}$ & Hyperparameters controlling the strength of regularization \\        
        $\gamma_1, \gamma_2, \gamma_3, \gamma_4$ & Hyperparameters to regularize $\bm{\Phi}^d_m$ \\
        $\sigma_{\bm{H}}, \sigma_{\bm{P}}$ & Hyperparameters that control the bandwidth of the kernel\\
        $\bm{\psi}_n^{ij} \in \mathbb{R}^{M_j, M_i}$ & Transformation of the data from state $i$ to state $j$ for channel $n$\\

        \hline
    \end{tabular}
   
    \label{tab:notations}
\end{table}

\begin{figure}[ht!]
\centering
    \includegraphics[width=0.52\textwidth]{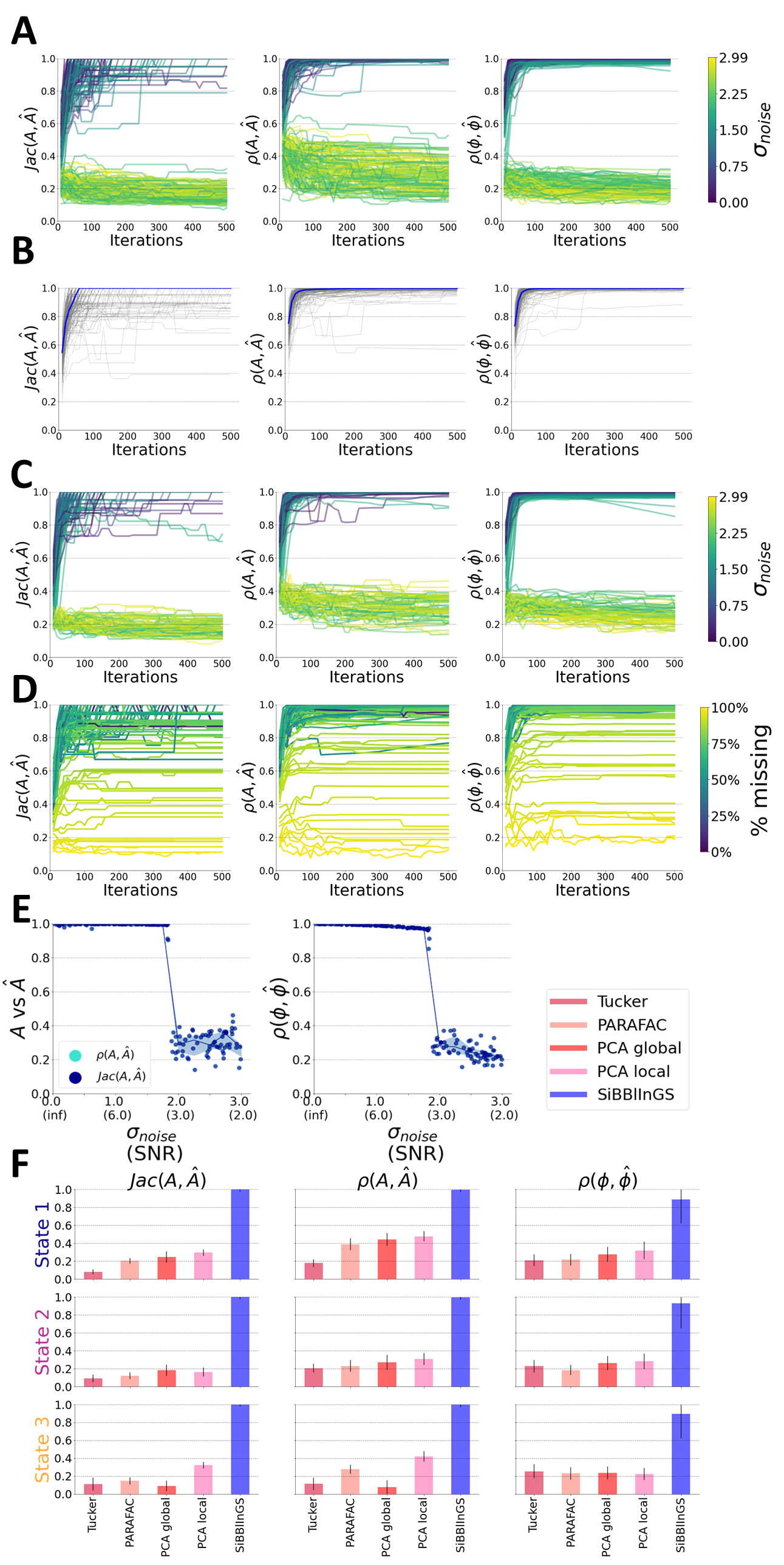}
  \caption{\textbf{Synthetic Data Results Robustness - cont.}
  \textbf{A}  Model performance under increasing levels of noise, along with random initializations, over the model training iterations. Color: increasing levels of missing samples. Left: Jaccard index between the recovered $\bm{A}$ and the ground true  $\bm{A}$. Middle: Correlation between the recovered $\bm{A}$ and the ground truth  $\bm{A}$. Right: Correlation between the recovered $\bm{\Phi}$ and the ground true $\bm{\Phi}$. 
  \textbf{B}  Model performance under random initializations (no noise), over the model training iterations. The blue curve is the median over all repeats.
  \textbf{C}  Model performance under increasing levels of noise only (fixed initializations). 
  \textbf{D} Model performance under increasing levels of missing samples, over the model training iterations. 
  \textbf{E} \sib Performance under increasing noise levels.
  \textbf{F} Comparison to other relevant methods, for each state individually (\sib in blue, other methods in pink to red colors). 
  }
   \label{fig:MSE_time}
\end{figure}

\begin{table}[h!]
\centering
 \caption{Assumptions and capabilities comparison between SiBBlInGS and other methods. }
\scalebox{0.8}{

\begin{tabular}{|c|c|c|c|c|c|c|c|c|c|}
\hline
\textbf{Method} & \textbf{SiBBlInGS} & \textbf{mTDR} & \textbf{PCA} & \textbf{Fast ICA} & \textbf{NMF} & \textbf{GPFA} & \textbf{SRM} & \textbf{HOSVD} & \textbf{PARAFAC} \\
\hline
Do not force orthogonality? & V & X & X & V & V & V & X & X & V \\
\hline
Sparse? & V & X & X & X & X & X & X & X & X \\
\hline
Flexible in time across states? & V	& X	& X	& X &	X&	X&	X&	V&	V \\
\hline
Support variations in BB across states? & V & X & X & X & X & X & X & X & X \\
\hline
Used for condition variability? & V & V & X & X & X & X & X & V & V\\
\hline
Works on tensors? & V & V & X & X & X & X & V & V & V\\
\hline
Consider both within & V & V & na & na & na & na & X & X & X \\
\& between states variability? & & & & & & & & & \\
\hline
Supports state-specific & V & V & na & na & na & na & V & X & X \\
emerging components? & & & & & & & & & \\
\hline
Works on non-consistent & V & X & na & na & na & na & X & X & X \\
data duration or & & & & & & & & & \\
sampling rates? & & & & & & & & & \\
\hline
Can prior knowledge & V & V & na & na & na & na & V & X & X \\
(labels) control state & & & & & & & & & \\
similarity? & & & & & & & & & \\
\hline
Ability to define both & V & V & na & na & na & na & X & X & X \\
state-specific and & & & & & & & & & \\
background components? & & & & & & & & & \\
\hline
Supports non-negative & V & X & X & X & V & X & X & X & V\\
decomposition? & & & & & & & & & \\
\hline
\end{tabular}}
\label{tab:comp_table}
\end{table}


\section{Further options for \texorpdfstring{$\bm{P}$}{P} computation}\label{sec:P_reco}

Here, we explore additional approaches for computing the state-similarity graph $\bm{P}$. These options take into account factors like data properties, single vs. multi-trial cases, variations in trial duration, and the desired approach (supervised or data-driven). 

\subsection{Supervised \texorpdfstring{$\bm{P}$}{P}}

In addition to the case presented in the paper, for sequential/ordered states, here we introduce the supervised version designed for categorical or similar-distanced states.
\subsubsection{Categorical or Similar-Distanced States}\label{sec:supervised_equal_case}
For cases where observation states are represented by categorical labels, and we expect a high degree of similarity between all possible pairs of states (i.e., no pair of labels is closer to each other than to another pair),
we can define the state similarity matrix $\bm{P}$ to ensure uniform values for all entries of distinct states, with larger values assigned to same-state entries located on the matrix diagonal.
$\bm{P}$ is then constructed as 
\begin{equation}
    \bm{P} = \bm{1} \otimes \bm{1}^T + c \bm{I},
\end{equation}
where $\bm{1} \otimes \bm{1}^T  \in \mathbb{R}^{D \times D}$ is a matrix of all ones, $\bm{I} \in \mathbb{R}^{D \times D}$ is the identity matrix, and $c$ is a weight that scales the strength of same-state similarity with respect to cross-state similarities. 

\subsection{Data-Driven \texorpdfstring{$\bm{P}$}{P}}\label{sec:unsup}
When prior knowledge about state similarity is uncertain or unavailable, SiBBlInGS also provides an unsupervised, data-driven approach to calculate $\bm{P}$ based on the distance between data points across states. 
Here we discuss the four options for constructing the matrix $\bm{P}$ in a data-driven manner, depending on the structure of the observations.

\subsubsection{Single-trial per-state with equal-length across states:}\label{kstep}
This case refers to the scenario of a single trial for each state ($M_d=1, \qquad \forall d = 1 \dots D$), where all cross-state trials have the same length ($T_1^d = T \, \qquad \forall d =1 \dots D$). Here, the similarity graph $\bm{P}$ is constructed as 
\begin{equation}
\bm{P}_{d,d'} = \exp\left(-{{||{\bm{Y}_1^d}- {\bm{Y}_1^{d'}}||_F^2}}/\sigma_{\bm{P}}^2\right), \label{P_data_driven_same_single}
\end{equation}

where $\sigma_{\bm{P}}$ controls the bandwidth of the kernel.

\subsubsection{Multiple trials per state, same trial duration}

In the case where all trials have the same temporal duration, the similarity matrix $\bm{P}$ is computed by evaluating the distance between the values of each pair of states, considering all trials within each state.
For this, we first find the transformation $\bm{\psi}^{ij}_{n} \in \mathbb{R}^{M_j\times M_i}$ between the observations of state $i$ to the observation of state $j$, by solving the Orthogonal Procrustes problem~\citep{golub2013matrix, gower2004dijksterhuis}. 
For this, let $\bm{Y}^{i\ast} \in \mathbb{R}^{M_i \times (T \times N)}$ be the matrix obtained by vertically concatenating the flattened observations from each trial ($m = 1 \dots M_i$) of state $i$. 
Then, the optimal transformation from the observations of state $i$ ($\bm{Y}^{i\ast} \in \mathbb{R}^{M_i \times (T \times N)}$) to the  observations of state $j$ ($\bm{Y}^{j\ast} \in \mathbb{R}^{M_j \times (T \times N)}$) will be 
\begin{gather}
    \widehat{\bm{\psi}}^{ij} = \underset{\bm{\psi}^{ij}}{\arg\min}\| \bm{\psi}^{ij} \bm{Y}^{i\ast} - \bm{Y}^{j\ast}\|_F^2, 
\end{gather}
where this mapping projects the multiple trials of state $i$ into the same space as of state $j$, via  {${\widetilde{\bm{Y}}^{i \ast} = \widehat{\bm{\psi}}^{ij} \bm{Y}^{i\ast}}$}. 
The state similarity matrix will thus be
\begin{equation}
    \bm{P}_{ij}  = \exp \left({-\|{\widetilde{\bm{Y}}^{i\ast}} - {\bm{Y}^{j\ast}}\|_F^2}/{{\sigma_p}^2}  \right),\label{P_for_multi_trial_same_dur}
\end{equation}
for all states $i\neq j = 1 \dots D$, where  $\sigma_p$ controls the kernel bandwidth.

\subsubsection{Single-trial per state, same duration} \label{single_trial_state_different_duration}

Further generalization of the state similarity computation requires addressing the case of trials with varying duration.  
When the observations correspond to the same process and their alignment using dynamic time warping is justifiable,
we can replace the Gaussian kernel measure with the Dynamic Time Warping (DTW) distance metric~\cite{berndt1994using}. In the case of a single trial for each state, the similarity metric becomes the average DTW distances over all channels, 
\begin{equation}
    \bm{P}_{ij}  = \exp{\left(-\frac{1}{N}\sum_{n=1}^N  DTW{(\bm{Y}_{n:}^i} , {\bm{Y}_{n:}^j)} \right).  }\label{P_data_driven_single}
\end{equation}

\subsubsection{Multiple trials per state, different duration} \label{use_of_DTW}
Similarly, for the multi-trial case we have
\begin{equation}
    \bm{P}_{ij} = \exp{\left(- \frac{1}{N} \sum_{n=1}^{N} \left( \frac{1}{M_i}
    \frac{1}{M_j}
    \sum_{m_j=1}^{M_j}
    \sum_{m_i=1}^{M_i}    
    \textrm{DTW}{ 
    \left( \left(\bm{Y}_{m_j}^j\right)_{n:} - \left(\bm{Y}_{m_i}^{i}\right)_{n:}    
    \right)} \right) 
    \right) }, \label{P_data_driven_multi}
\end{equation}

where, $\textrm{DTW}$, is, as before, the Dynamic Time Wrapping~\cite{berndt1994using} operator, applied on the activity of the $n$-th channel in both states. 
It is crucial to note that this approach operates under the assumption that the trials being compared depict similar processes, and hence aligning them using DTW is a valid assumption.

\section{Channel-similarity kernel (\texorpdfstring{$\bm{H}$}{H})\textemdash generation and processing}\label{sec:channel_H}

The kernel post-processing involves several steps. First, we construct the kernel ${\widetilde{\bm{H}}}^d$ for each state $d = 1\dots D$, as described in the main text.
To incorporate similarities between each possible pair of states $d' \neq d$, where $d, d' = 1 \dots D$, we perform a weighted average of each $\widetilde{\bm{H}}^d$ with the kernels of all other states, using $\bm{P}_{d:}$ for the weights, as it quantifies the similarity between state $d$ and all other states: 
$\bm{H}^d = \sum_{d'=1}^{D} \bm{P}_{dd'} \widetilde{\bm{H}}^{d'}$.
Then, to promote a more robust algorithm, we only retain the $k$ highest values (i.e., k-Nearest Neighbors; kNN) in each row, while the rest are set to zero. The value of $k$ is a model hyperparameter, and depends on the desired BB size. 
We then symmetrize each state's kernel by calculating $\bm{H}^d \leftarrow \frac{1}{2} \left(\bm{H}^d + (\bm{H}^d)^T\right)$ for all $d = 1 \dots D$.
Finally, the kernel is row-normalized so that each row sums to one, as follows: Let $\Lambda^d$ be a diagonal matrix with elements representing the row sums of $\bm{H}^d$, i.e., $\bm{\Lambda}^d_{ii} = \sum_{n=1}^N \bm{H}_{i,n}^d $. The final normalized channel similarity kernel is obtained as $\bm{H}_\textrm{final}^d = (\bm{\Lambda}^d)^{-1} \bm{H}^d$.

\section{Solving \texorpdfstring{$\bm{\Phi}$}{Phi} in practice}\label{solve_phi}

In Section~\ref{update_over_phi}, the model updates the temporal traces dictionary $\bm{\phi} = \bm{\Phi}_m^d$ for all $m = 1 \dots M_d$, $d = 1 \dots D$ using an extended least squares for each time point $t$, i.e., 
\begin{equation}\label{eqn:update_phi_practice} 
    \widetilde{\bm{\phi}}_{[t,:]} = \arg\min_{\bm{\phi}_{[t,:]}} \Vert\widetilde{\bm{Y}}_{m_{[:,t]}}^d -\widetilde{\bm{M}}\bm{\phi}_{[t,:]} \Vert_2^2, 
\end{equation}
where $\bm{\phi}_{[t,:]} \in \mathbb{R}^{p}$ is the dictionary at time $t$,

\begin{equation}
\widetilde{\bm{Y}}_{m_{[:,t]}}^d = \begin{bmatrix}  \bm{Y}_{m_{[:,t]}}^d \\ \ \bm{[0]}_{p\times 1} \\ \gamma_2 \bm{\phi}_{[t,:]}^{(iter-1)}  + \gamma_3 \bm{\phi}_{[(t-1),:]} \end{bmatrix}
,\quad \text{ and } \quad         
\widetilde{\bm{M}} = 
\begin{bmatrix} \bm{A}^d \\  
\gamma_4 ( \frac{[1]_{p\times p}}{p}  - \bm{I}_{p\times p}) \circ \sqrt{D} \\ 
({\gamma_1  + \gamma_2 + \gamma_3}) \bm{I}_{p\times p}  \end{bmatrix},
\nonumber\
\end{equation}
with all parameters being the same as those defined in Section~\ref{update_over_phi} of the main text and $\circ$ denotes element-wise multiplication.
Here, $[0]_{p \times 1} \in \mathbb{R}^{p\times 1}$ represents a column vector of zeros,  $[1]_{p\times p}$ represents a square matrix of ones with dimensions $p \times p$, and $\bm{Y}_{m_{[:,t]}}^d \in \mathbb{R}^{N}$ denotes the measurement in the $m$-th trial of state $d$ at time $t$.

\section{Model Complexity}\label{sec:model_complexity}
SiBBIInGS relies on 4 main computational steps: 

\textbf{Channel Graph Construction:} This operation, performed once for all $N$ channels of every state $d = 1 \dots D$, generates a channel graph $\bm{H}^d \in \mathbb{R}^{N \times N}$ for each state $d \in[1, D]$ by concatenating within-state trials $1 \dots M_d$ horizontally, resulting in a $N \times \sum_{m=1}^{M_d} T_m^d$ matrix. 
For simplicity, let $\widetilde{T} = \sum_{m=1}^{M_d} T_m^d$.
The computational complexity of calculating the pairwise similarities of this concatenated matrix for all $D$ states is thus $\mathcal{O}\left(D \widetilde{T}^2 N(N-1)\right)$. 

For the k-threshold step (\ref{kstep}), that involves keeping only the $k$ largest values in each row while setting the other values to zero\textemdash the complexity will be $\mathcal{O}\left(\widetilde{T} \log k\right)$ per row for a total computational complexity of  $\mathcal{O}\left(DN \widetilde{T} \log k\right)$   for $N$ rows and $D$ states.

\textbf{State Graph Construction:} This is a one-time operation that involves calculating the pairwise similarities between each pair of states. For simplicity, if we assume the case of user-defined scalar labels, and as in this case there are $D$ states (and accordingly $D$ labels), the computation includes $D(D-1)) / 2$ pairwise distances for $\mathcal{O}\left(D^2\right)$.

\textbf{BB Inference (Eq.~\eqref{eqn:AUpdate1}):} This iterative step involves per-channel re-weighted $\ell_1$ optimization. If the computational complexity of a weighted $\ell_1$ is denoted as $\mathcal{C}$, then the computational complexity of the re-Weighted $\ell_1$ Graph Filtering is $N L \mathcal{C}+L N k$, where $N$ is the number of channels, $L$ is the number of iterations for the RWLF procedure, and $k$ is the number of nearest neighbors in the graph. 
For the last term in Eq.~\eqref{eqn:AUpdate1}, there are $p^2$ multiplicative operations involving the vector $\nu$ and the difference in BBs, arising from the $\ell_2^2$ norm. Additionally, there is an additional multiplication step involving  $\bm{P}_{dd'}$. For each state $d$, this calculation repeats itself $D - 1$ times (for all $d' \neq d$). This process is carried out for every $d = 1 \dots D$. In total, these multiplicative operations sum up to $\left(p^2+1\right) D(D-1)$, resulting in a computational complexity of $\mathcal{O}\left(D^2 p^2\right)$.

\textbf{Optimization for $\bm{\phi}$}:
This step refers to the least-squares problem presented in Eq.~\eqref{eqn:update_phi_practice} in Appendix~\ref{solve_phi}.
If a non-negative constraint is applied, SiBBlInGS uses scipy's ``nnls'' for solving $ \widetilde{\bm{\phi}}_{[t,:]} = \arg\min_{\bm{\phi}_{[t,:]}} \Vert\widetilde{\bm{Y}}_{m_{[:,t]}}^d -\widetilde{\bm{M}}\bm{\phi}_{[t,:]} \Vert_2^2$, where ${\widetilde{\bm{Y}}^d_m \in R^{(N+2 p) \times T_m^d}}$ and $\bm{M} \in R^{(N+2 p) \times p}$. This results in complexity of $\mathcal{O}\left(p (N+2p)^2 F \max{(T_m^d)}\right)$, where $F$ is the number of nnls iterations. Without non-negativity constraint, this problem is a least squares problem with a complexity of $\mathcal{O}\left(\max{(T_m^d)} p^2 (2p+N))\right)$. Potential complexity reduction options include parallelizing RWL1 optimizations per channel, using efficient kNN or approximate kNN search for constructing kNN graphs instead of full graphs, and employing dimensionality reduction techniques to expedite nearest neighbor searches.

\section{Data and Code Availability}
The code employed in this study is available on \href{https://github.com/NogaMudrik/SiBBlInGS}{https://github.com/NogaMudrik/SiBBlInGS}. The data used in this study are publicly available and cited within the paper.


\section{General Experimental Details}
All experiments and code were developed and executed using Python version 3.10.4 and are compatible with standard desktop machines. 

\section{Synthetic Data\textemdash Additional Information}
\begin{figure}[t!]
\centering
\includegraphics[width=1\textwidth]{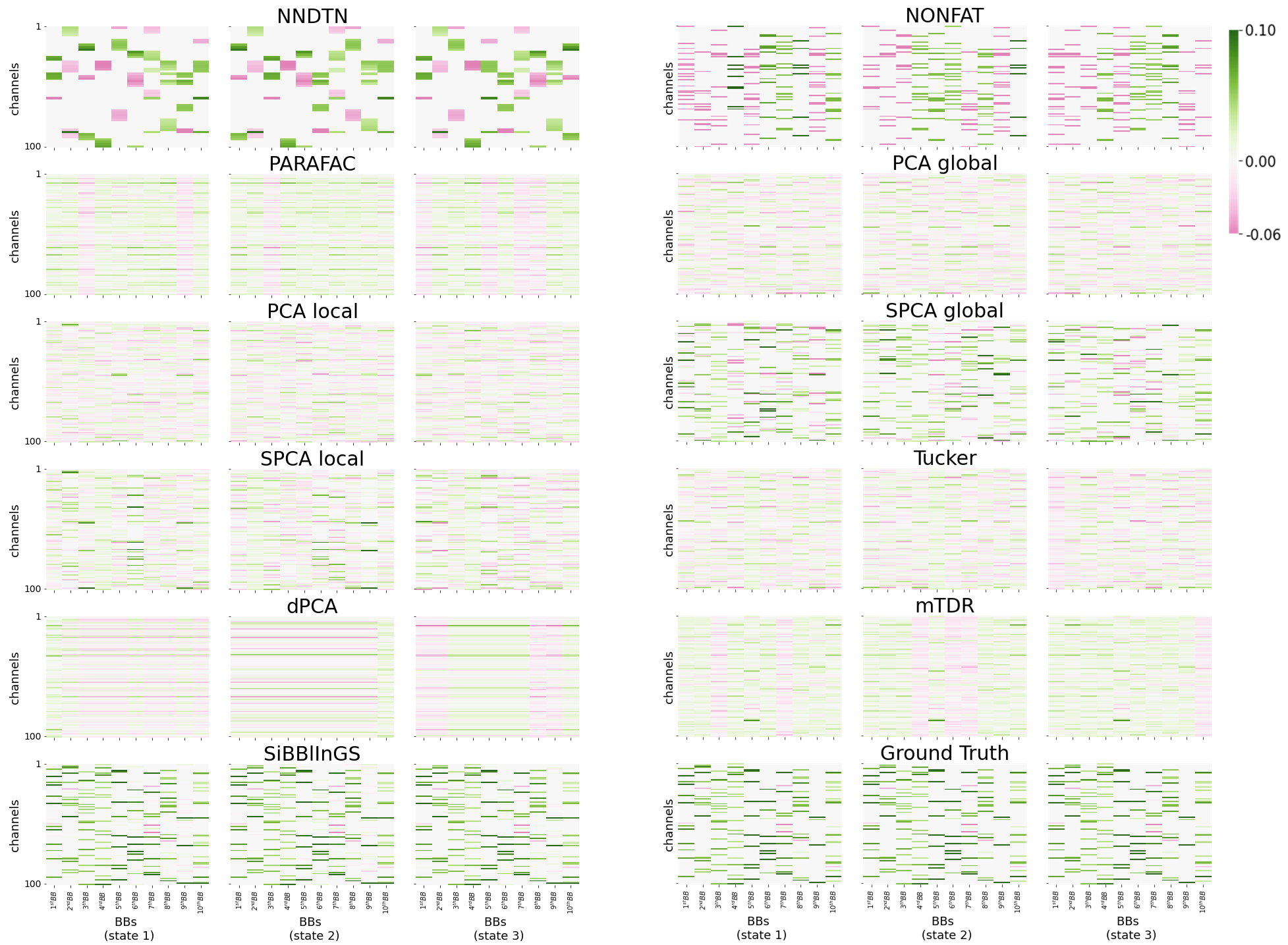}
\caption{\textbf{BBs identified by different methods}. BBs identified by SiBBlInGS are compared with those from other methods, including PARAFAC, Tucker, PCA (global and local), Sparse PCA (global and local), demixed PCA~\cite{kobak2016demixed}, mTDR~\cite{aoi2018model}, and Gaussian-process-based methods~\cite{wang2022nonparametric}. The identified BBs were reordered to best match the ground truth BBs' temporal traces through maximum correlations. A subsequent hard-thresholding step was applied to achieve sparsity, aligning with the sparsity level with of the ground truth components. The BBs were normalized to have an absolute sum of 1 each for visualization purposes. 
}
\label{fig:BBs_themselves}
\end{figure}

\begin{figure}[t]
\centerline{\includegraphics[width=0.8\columnwidth]{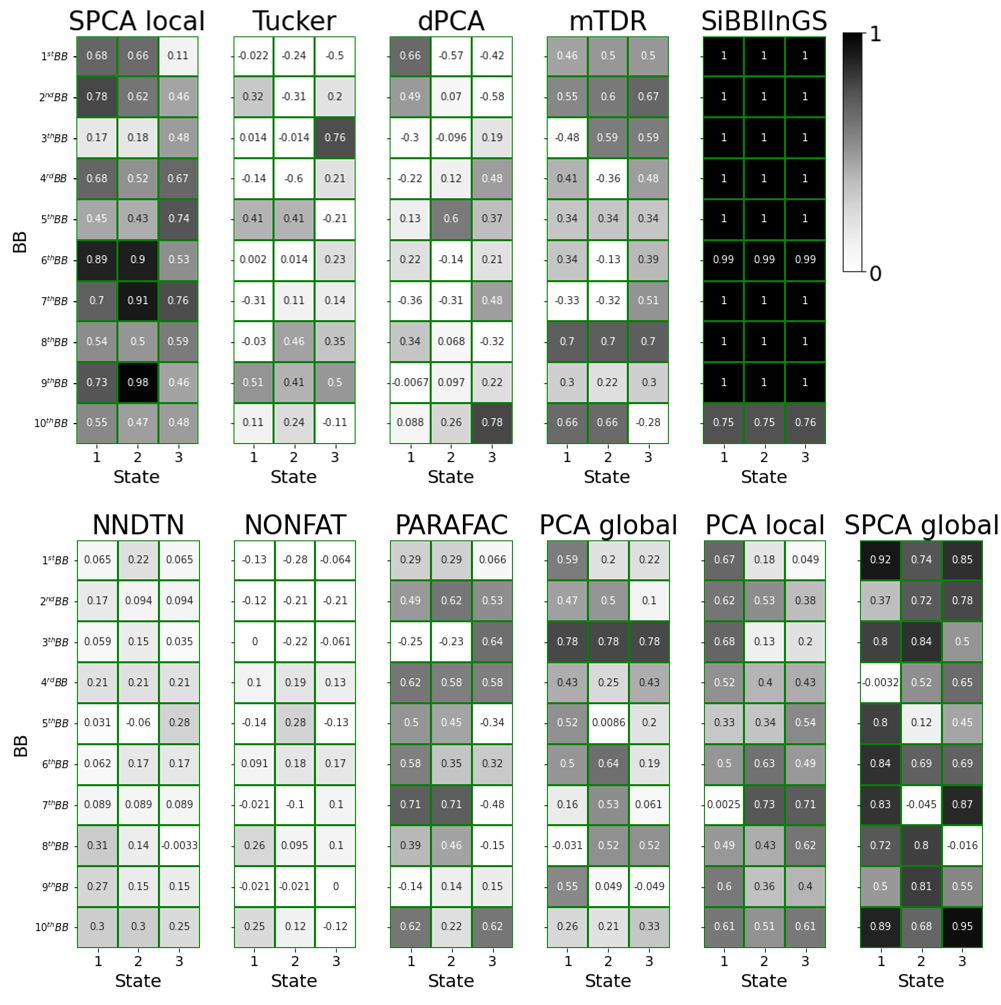}}
\captionsetup{format=hang}
\caption{Correlations between BBs identified by different methods and ground truth BBs for each state and BB number.}
\label{icml-historical}
\end{figure}

\subsection{Synthetic Generation Details}

We initiated the synthetic data generation process by setting the number of channels to $N = 100$ and the maximum number of BBs to $p = 10$. We further defined the number of states as $D = 3$ and determined the number of time points in each observation to be $T^d = 300$, where $d$ represents the state index (here $d\in\{1,2,3\}$). We defined the number of trials for each state as one, i.e., $M_d = 1$ for $d = 1,2,3$.

We first initialized a  ``general'' BB matrix ($\bm{A}$) as the initial structure, which will later undergo minor modifications for each state.

For each state $d$, we generated the time-traces $\bm{\Phi}^d$
via a linear combination of $15$ trigonometric signals, such that the temporal trace of the $j$-th BB was defined as $\bm{\Phi}_{:j}^d = \sum_{i=1}^{15} c_i f_i(\text{freq}_i*x)$ where $x$ is an array of $T = 300$ time points ($x = [1,\dots, 300]$), $\text{freq}_i$ is a random scaling factor sampled uniformly on $[0, 5]$, $f$ refers to a random choice between the sine and cosine functions (with probability 1/2 for each), and the sign ($c_i$) was flipped ($+1$ or $-1$) with a probability of 1/2.

During the data generation process, we incorporated checks and updates to $\bm{A}$ and $\bm{\Phi}$ to ensure that the BBs and their corresponding time traces are neither overly correlated nor orthogonal, are not a simple function of the states labels, and that different BBs exhibit comparable levels of contributions. This iterative process involving the checks persisted until no further modifications were required.


The first check aimed to ensure that the temporal traces of at least two BBs across all states were not strongly correlated with the state label vector ($[1, 2, 3]$) at each time point. Specifically, we examined whether the temporal traces of a $j$-th BB across all states ($\Phi_{tj}^1, \Phi_{tj}^2, \Phi_{tj}^3$) exhibited high correlation with the state label vector at each time point. This check was important to avoid an oversimplification of the problem by preventing the temporal traces from being solely influenced by the state labels.
To perform this check, we calculated the average correlation between the temporal traces and the state labels ($[1, 2, 3]$) at each time point. If the average correlation over time exceeded a predetermined threshold of 0.6, we introduced additional variability in the time traces of the  BB that exhibited a high correlation with the labels. This was achieved by adding five randomly generated trigonometric functions to the corresponding BB. These additional functions were generated in the same manner as the original data (with $\bm{\Phi}_{:j}^d = \sum_{i=1}^{5} c_i \bm{f}_i(\text{freq}_i \cdot x)$). 

The second check ensured that the time traces were not highly correlated with each other and effectively represented separate functions.
If the correlation coefficient between any pair of temporal traces of different BBs  exceeded a threshold of $\rho = 0.6$, the correlated traces were perturbed by adding zero-mean Gaussian noise with a standard deviation of $\sigma = 0.02$.

Next, we ensured that the BBs represented distinct components by verifying that they were not highly correlated with each other.
Specifically, if the correlation coefficient between a pair of BBs ($\bm{A}_{:j}$, $\bm{A}_{:i}$ for $j,i = 1 \dots 10$) within a state exceeded the threshold $\rho = 0.6$, each BB in the highly-correlated pair was randomly permuted to ensure their distinctiveness.

To prevent any hierarchical distinction or disparity in BB contributions and distinguish our approach from order-based methods like PCA or SVD, we evaluated each BB's contribution by measuring the increase in error when exclusively using that BB for reconstruction. Specifically, for the $j$-th BB, we calculated its contribution as 
$
\text{contribution}_j = -\|\bm{Y} - \bm{A}_{:j} \otimes \bm{\Phi}_{:j}\|_F
$, where $\otimes$ denotes the outer product.
Then, we compared the contributions between every pair of BBs within the same state. 
If the contribution difference between any pair of BBs exceeded a predetermined threshold of $10$, both BBs in the pair were perturbed with random normal noise. Subsequently, a hard-thresholding operation was applied to ensure that the desired cardinality was maintained.

To introduce slight variability also in the BBs' structures across states (in addition to the temporal variability), the general basis matrix $\bm{A}$ underwent modifications for each of the states. In each state and for each BB, a random selection of 0 to 2 non-zero elements from the corresponding BB in the original $\bm{A}$ matrix were set to zero, effectively introducing missing channels as differences between states, such that $\bm{A}^d$ is the updated $\bm{A}$ modified for state $d$.
Finally, the data was reconstructed using $\bm{Y}^d = \bm{A}^d (\bm{\Phi}^d)^T$ for each state $d = 1,2,3$.

\subsection{Experimental details to the Synthetic data}\label{sec:synth_gen}

We applied SiBBLInGS to the synthetic data with $p = 10$ components and a maximum number of $10^3$ iterations, while in practice about $50$ iterations were enough to converge (see Fig.~\ref{fig:MSE_time}). The parameters for the $
\lambda$ update in Eq.~\eqref{eqn:updateLambda} were $\epsilon = 0.01$, $\beta  = 0.09$, and $w_\text{graph} = 1$.
For the regularization of $\Phi$ (Eq.~\eqref{eqn:update_phi}), the parameters used were $\gamma_1 = 0.1$, $\gamma_2 = 0.1$, $\gamma_3 = 0$, and $\gamma_4 = 0.0001$. 
$\nu$ was set to be a vector of ones with length $p = 10$.
The number of repeats for the update of $\bm{A}$ within an iteration, for each state, was set to $2$. 
The number of neighbors used in the channel graph reconstruction 
($\bm{H}^d$)
was $k = 25$.
The python scikit-learn's~\cite{scikit-learn} LASSO solver was used for updating $\bm{A}$ in each iteration. 
This synthetic demonstration used the supervised case for building $\bm{P}$, where $\bm{P}$ was defined assuming similar similarity levels between each pair of states, by defining $\bm{P} = \bm{1} \otimes \bm{1}^T \in \mathbb{R}^{3\times 3}$ (the case described in  App.~\ref{sec:supervised_equal_case}, with $c = 1$).

\subsection{Jaccard index calculation}
In Figure~\ref{fig:intro_fig}C, we computed the Jaccard similarity index between the identified BBs by SiBBlInGS and the ground truth BBs. To obtain this measure, we first rearranged the BBs based on the correlation of their temporal traces with the ground truth traces (since the method is invariant to the order of the BBs). Then, we nullified the 15 lower percentiles of the $\widehat{\bm{A}}$ matrix, which correspond to values very close to zero. Finally, we compared the modified identified BBs to the ground truth BBs using the ``jaccard\_score'' function from the sklearn library~\cite{scikit-learn}.

\section{Google Trends\textemdash Additional Information}
\subsection{Trends data acquisition and pre-processing}\label{sec:trends_pre}
\begin{figure}[t]
\centering
\includegraphics[width=1\textwidth]{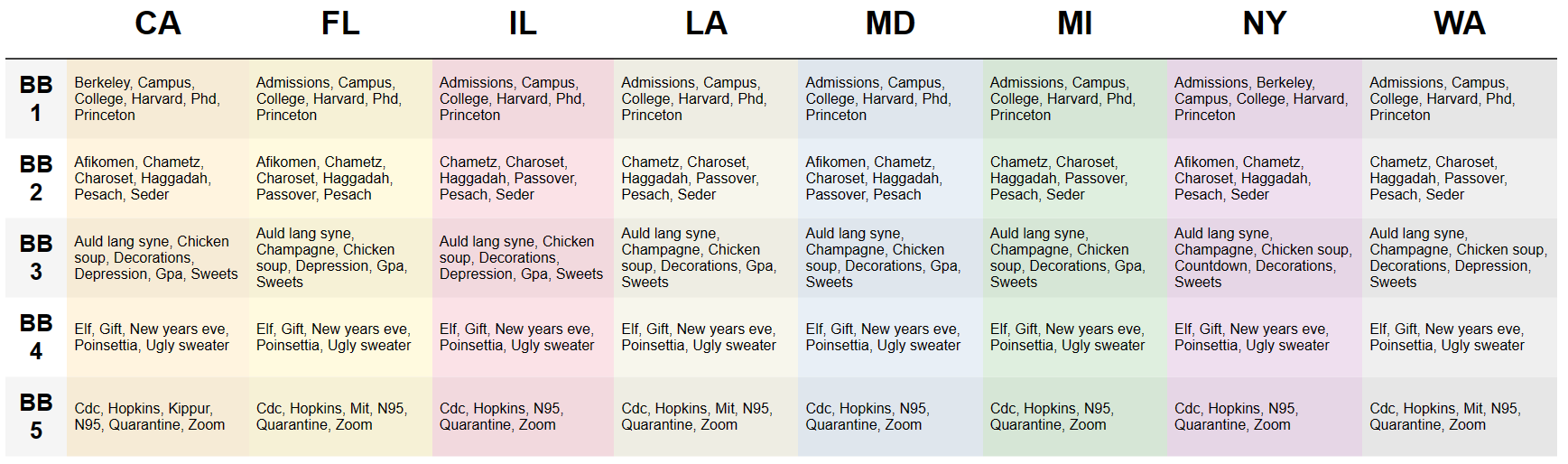}
\caption{\textbf{Table of clustered words for the Google Trends experiment}}
\label{fig:trends_table}
\end{figure}

The acquisition and pre-processing of Google Trends data involved manually downloading the data from April 1, 2010, to November 27, 2022, for each of the selected states:
California (CA), Maryland (MD), Michigan (MI), New York (NY), Illinois (IL), Louisiana (LA), Florida (FL), and Washington (WA), directly from the Google Trends platform~\citep{GoogleTrends}.
The full list of terms (including their clusters as SiBBlInGS recovered)
is presented in Figure~\ref{fig:trends_table}.
To ensure coverage of search patterns, the data was downloaded by examining each query in all capitalization formats, including uppercase, lowercase, and mixed case.

\begin{figure}
\centering
\includegraphics[width=1\textwidth]{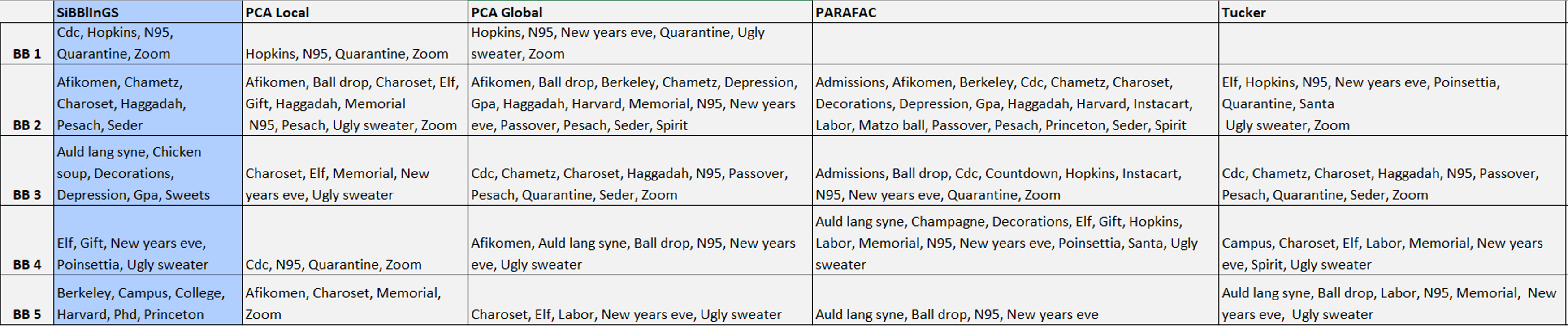}
\caption{\textbf{Comparison of The Google Trends Results to Other Methods with 5 BBs for CA:} Comparison to other methods, each applied with $p = 5$ BBs, yielded less interpretable BBs. For example, SiBBlInGS discerns theme-specific BBs (e.g., 'Covid' and 'University'), while other methods produce more blended compositions. Empty cells for PARAFAC and Tucker indicate that those BBs remained empty.
}
\label{fig:trends_comparison}
\end{figure}

\begin{figure}
\centering
\includegraphics[width=0.8\textwidth]{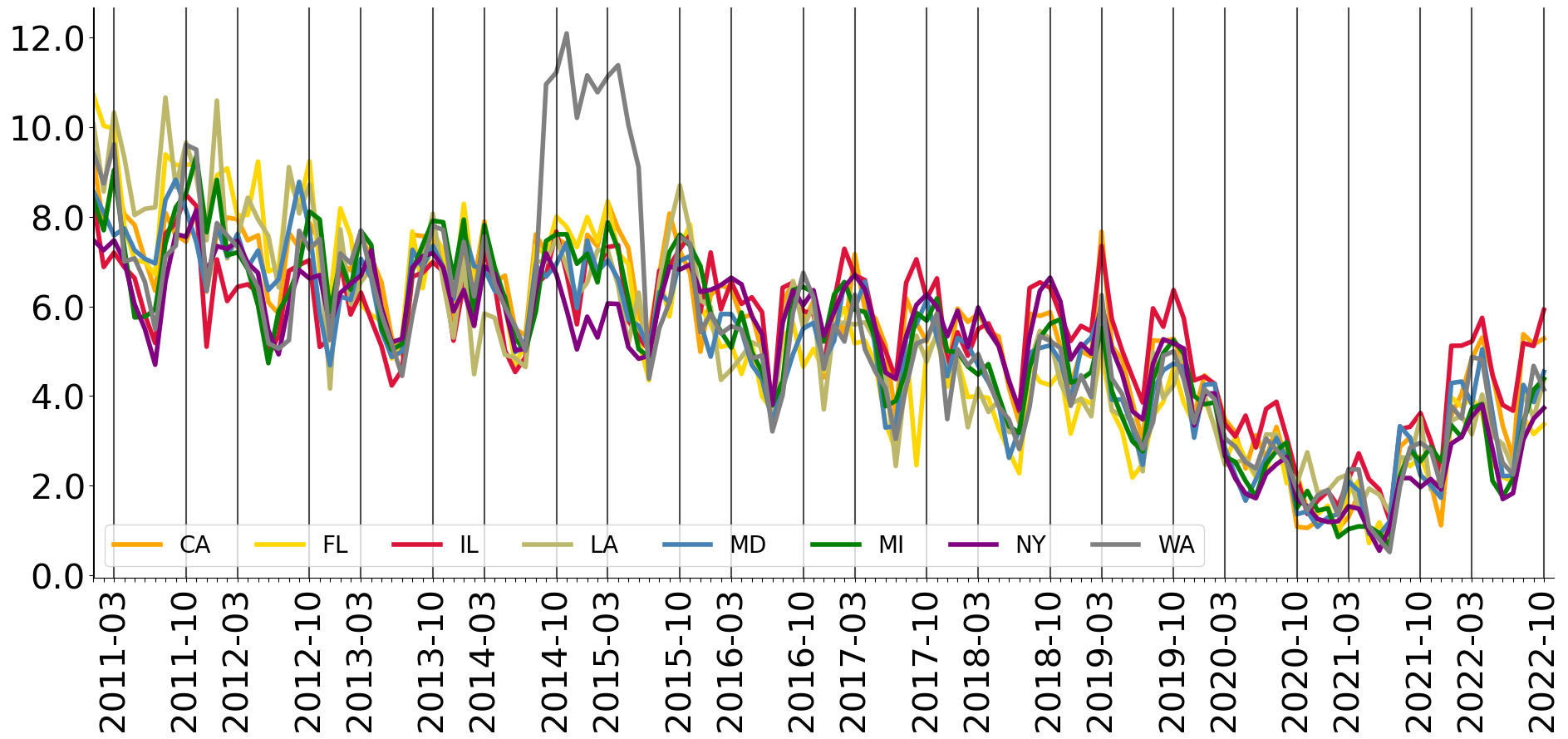}
\caption{Temporal traces of one of the BBs SiBBlInGS identified, which includes college admission terms, show bi-yearly peaks around March and October, aligning with key milestones in the US college admissions process. Additionally, a decrease in online interest in the college BB is observed during the COVID-19 pandemic.}
\label{fig:trends_college}
\end{figure}

The data (in CSV format) was processed using  the ``pandas'' library in Python~\citep{reback2020pandas, mckinney-proc-scipy-2010} and keeping only the relevant information from January 2011 to October 2022, inclusively. 
We conducted a verification  to ensure the absence of NaN (null) values for each term in every selected state. 
To pre-process each term, we implemented a two-step normalization procedure. First, the values within the chosen date range were scaled to a maximum value of 100. This step ensured that the magnitude of each term's fluctuations remained within a consistent range. Next, the values for each term were divided by the sum of values across all dates and then multiplied by 100, resulting in an adjusted scale where the area under the curve for each term equaled 100. This normalization procedure  accounted for potential variations in the frequency and magnitude of term occurrences, enabling fair comparisons across different terms.
By applying these pre-processing steps, we aimed to mitigate the influence of isolated spikes or localized peaks that could distort the overall patterns and trends observed in the data.
Since the focus of this processing was on assessing the relative contribution of a term within a BB rather than comparing the overall amplitude and mean of the term across states, factors such as population size and other characteristics of each state were not taken into consideration. 

\subsection{Experimental details for Google trends}

We ran the Trends experiment with $p = 5$ BBs, and applied non-negativity constraints to both the BB components and their temporal traces.  
The $\bm{\lambda}$'s parameters in Equation~\eqref{eqn:AUpdate1} included $\epsilon = 9.2$, $\beta = 0.01$, and $w_\text{graph} = 35$.
For the regularization of $\bm{\Phi}$ in Equation~\eqref{eqn:update_phi}, we used the parameters $\gamma_1 = 0$, $\gamma_2 = 0$, $\gamma_3 = 0.05$, $\gamma_4 = 0.55$. 
The trends example used the data-driven version for studying $\bm{P}$, and we set $\bm{\nu}$ to be a vector of ones with length $p = 5$.

During each iteration, $\bm{A}$ underwent two updates within each state $d$.
The number of neighbors we used in the channel graph reconstruction was $k = 4$.
We used the PyLops package in Python, along with the SPGL1 solver~\citep{ravasi2020pylops} to update $A$ in each iteration. With respect to SPGL1 parameters (as described in~\citep{ravasi2020pylops}), we set the initial value of the parameter $\tau$ to $0.12$, and a multiplicative decay factor of $0.999$ was applied to it at each iteration.
We note here that SPGL1 solves a Lagrangian variation of the original Lasso problem, where, i.e., it bounds the $\ell_1$ norm of the selected BB to be smaller than $\tau$, rather than adding the $\ell_1$ regularization to the cost~\citep{vandenberg2008probing, ravasi2020pylops}.

\subsection{Temporal traces of college BB}\label{sec:college}
The temporal traces of the BBs that relate to college admission, as identified by SiBBlInGS, exhibit distinct bi-yearly peaks, with notable increases in activity around March and October, along with a clear decrease between March to next October (Fig.~\ref{fig:trends_college}). These peaks align with key periods in the US college admissions cycle, including application submission and admission decision releases.
Particularly, around the end of March, many colleges and universities in the US release their regular admission decisions, prompting increased population interest. 
Similarly, October marks the time when prospective students typically start showing increased interest in applying to colleges, as many colleges have early application deadlines that fall in late October or early November.
The bi-yearly peaks pattern in March and October thus reflects the concentrated periods of activity and anticipation within the college admissions process. 
External factors such as the COVID-19 pandemic can also influence the timing and dynamics of the college admissions process, as we observe by the decrease in the college BB activity during the pandemic period (Fig. ~\ref{fig:trends_college}). 

\subsection{Temporal traces of ``Passover'' BB}\label{sec:passover}
\begin{figure}[t]
\includegraphics[width=1\textwidth]{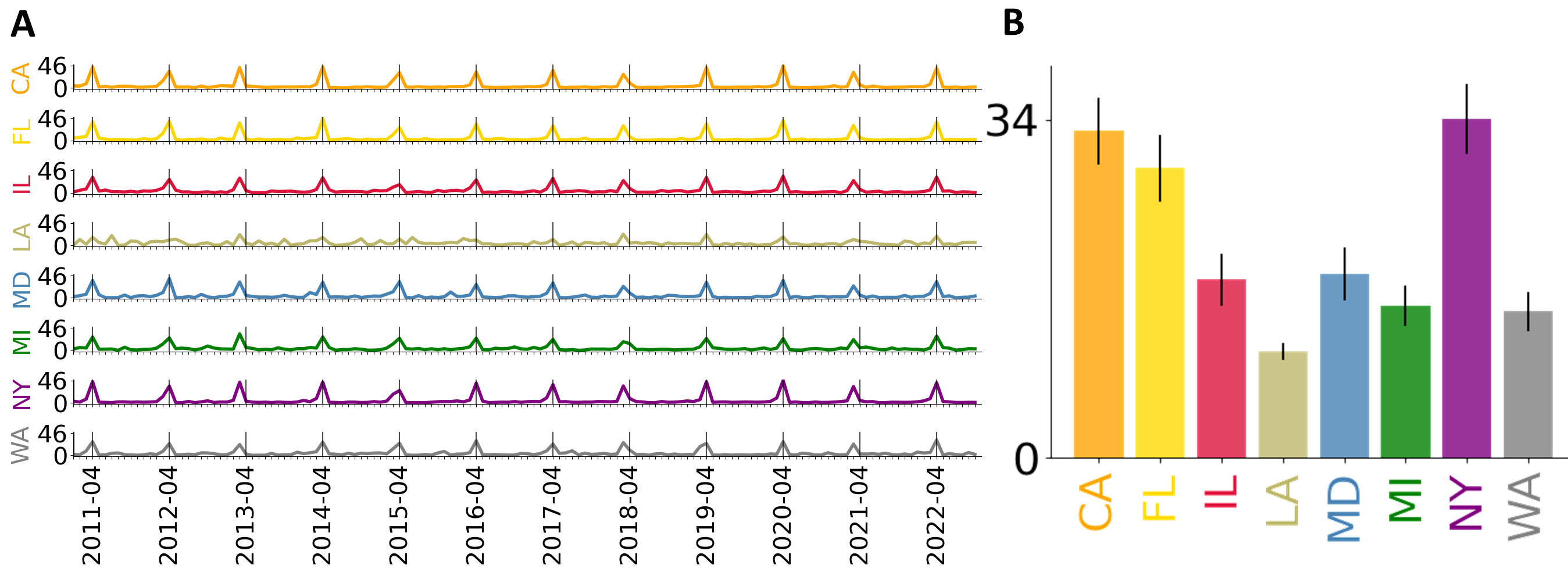}
\caption{\textbf{Temporal trace of Passover BB}. The Passover's BB patterns, as SiBBlInGS found,  show an alignment with the percentage of Jewish population in different states. 
\textbf{A} Temporal traces of the Passover BB for each state. Vertical black lines indicate the month of April, when Passover is usually celebrated.  \textbf{B} The mean and standard error of peak values for each state.}
\label{fig:trends_passover}
\end{figure}

SiBBInGS identified a ``Passover'' BB, characterized by temporal traces that show a clear alignment with the timing of Passover, which usually occurs around April. The time traces demonstrate a prominent peak in states with higher Jewish population percentages, like CA, FL,  and NY, as presented by the average peak value in Figure~\ref{fig:trends_passover}B plotted for the different states.
The peaks detection (in Fig.~\ref{fig:trends_passover}) was done using scipy's~\citep{scipy} ``find\_peaks'' function with a threshold of 4.

\section{Neural Data\textemdash Additional Information}

\subsection{Neural Data Pre-Processing}

In this experiment we used the neural data collected from Brodmann's area 2 of the somatosensory cortex in a monkey performing a reaching-out movement experiment from Chowdhury et al. \cite{chowdhury2022dandiarchive, chowdhury2020area}. 
While the original dataset includes data both under perturbed and unperturbed conditions, here, for simplicity, we used only unperturbed trials. 
We followed the processing instructions provided by Neural Latents Benchmark~\cite{PeiYe2021NeuralLatents} to extract the neural information and align the trials. 
The original neural data consisted of spike indicators per neuron, which we further processed to approximate spike rates by convolving them with a Gaussian kernel.

For each of the 8 angles, we randomly selected 18 trials, resulting in a total of 144 data matrices. The states were defined as the angles, and for learning the supervised $\bm{P}$, we used as labels the x-y coordinates of each angle in a circle with a radius of 1 (i.e., sine and cosine projections). 

\subsection{Experimental details for the neural data experiment}

We ran SiBBLInGS on the reaching-out dataset with $p = 4$ BBs.  
The $\bm{\lambda}$'s parameters used were $\epsilon = 2.1$, $\beta = 0.03$, and $w_\text{graph} = 10.1$. 
For the regularization of $\bm{\Phi}$ we used: $\gamma_1 = 0.001$, $\gamma_2 = 0.001$, $\gamma_3 = 0.1$, and $\gamma_4 = 0.3$ and we set $\bm{\nu}$ to be a vector of length $p = 4$ with $\nu_1$ = 0.8  (to allow more flexibility in the first BB), and $\nu_k = 1$ for $k=2,3,4$. 
For the neural data, we used the supervised version of $\bm{P}$, where the $x-y$ coordinates are used as the labels for calculating $\bm{P}$. 
During each iteration, $\bm{A}$ underwent two updates within each state.
We chose $k = 7$ neighbors for the channel graph reconstruction, and used 
Python scikit-learn's~\citep{scikit-learn} LASSO solver for the update of $\bm{A}$.

\subsection{State prediction using temporal traces}\label{sec:predict}

We used the identified temporal traces $\bm{\Phi}$ to predict the state (hand direction). The dimensionality of each state's temporal activity $\bm{\Phi}^d$ was reduced to a vector of length $p \times 4 = 16$ using PCA with 4 components.
A k-fold cross-validation classification approach with $k=4$ folds was used in a multi-class logistic regression model with multinomial loss (trained on 3 folds and used to predict the labels of the remaining fold). This process was repeated for each fold, and the results were averaged.
The confusion matrix and accuracy scores for each state (angle), are shown in Figure~\ref{fig:neuro_fig2}C and in Figure~\ref{fig:neuro_supp_fig}F.

\subsection{Computation of  $\rho_{\text{within}/\text{between}}$}\label{sec:within_bet}

To compute the correlation for the ``within'' state case, a random bootstrap approach was used, such that, for each state, we randomly selected 100 combinations of temporal trace pairs corresponding to the same BB from different random trials within the same state. We computed the correlations between these temporal trace pairs, and averaged the result over all 100 bootstrapped pairs to obtain the average correlation.

\begin{figure}
\includegraphics[width=0.95\textwidth]{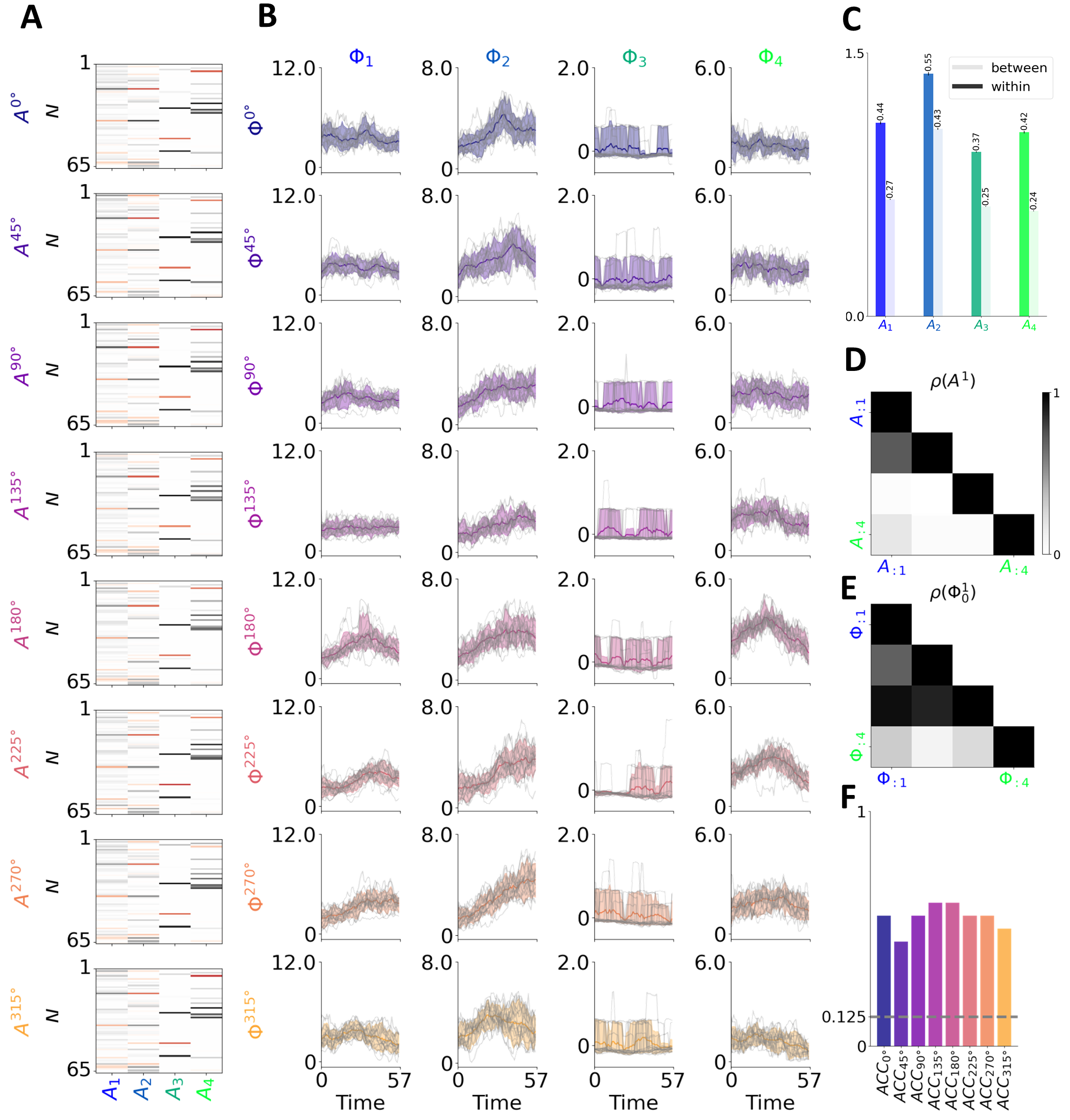}
\caption{\textbf{Additional Figures for the Neural Recordings Experiment}. \textbf{A} The identified BBs for the different states. While there is clear consistency, slight modifications can be observed across states, capturing the natural variability in neural ensembles corresponding to different tasks. 
\textbf{B} Temporal traces of the identified BBs, shown with a 90\% confidence interval (background color), and all trials are plotted in light gray. The color corresponds to the state color used in Figure~\ref{fig:neuro_fig2}. We observe adaptation over the states as well as differences between the temporal traces of BBs within a given state. The third BB exhibits significantly lower activity compared to the others (see also Figure~\ref{fig:neuro_fig2}), suggesting that it might capture general background trends or noise.  
\textbf{C} Within and between temporal trace correlations (averaged over 100 bootstrapped examples) with standard error, colored according to the BB color (as used in  Fig.~\ref{fig:neuro_fig2}), and transparency representing the strength of the between (opaque) and within (less opaque) correlations. 
\textbf{D} Example of the correlations between each pair of BBs within the $1$-st state ($0$\textdegree). This shows that while some BBs are orthogonal, others are not. 
\textbf{E} Example of within-state correlations between each pair of temporal traces of the BBs within the $1$-st trial of the $1$-st state ($0$\textdegree), showing that the temporal traces are neither orthogonal nor overly correlated. 
\textbf{F} Accuracy in predicting the state using only the temporal traces of that state as input (colored by the state color). While the random accuracy would be $1/\textrm{length(labels)} = \frac{1}{8} = 0.125$, the achieved accuracies are significantly higher for all states.  }
\label{fig:neuro_supp_fig}
\end{figure}

Similarly, for the ``between'' states case, we repeated this procedure for pairs of trials from distinct states. Particularly, we selected 100 random bootstrapped combinations of pairs of the same BB from trials of different states.
In Figure~\ref{fig:neuro_supp_fig}C, the average correlations are shown for each BB. The ratio depicted in Figure~\ref{fig:neuro_fig2}E represents the ratio between the averages of the ``within'' and ``between'' state correlations.

\section{Epilepsy\textemdash Additional Information}\label{sec:epi_compare}
\subsection{Data Characteristics and Pre-processing for SiBBlInGS Analysis}
The Epilepsy EEG experiment in this paper is based on data kindly shared publicly in~\citep{handa2021open}.

The data consist of EEG recordings obtained from six patients diagnosed with focal epilepsy, who were undergoing presurgical evaluation. As part of this evaluation, patients temporarily discontinued their anti-seizure medications to facilitate the recording of habitual seizures. The data collection period spanned from January 2014 to July 2015.
These seizures manifest different patients, seizure types, ictal onset zones, and durations.

The EEG data, as described by~\citep{handa2021open}, were recorded using a standard 21 scalp electrodes setup, following the 10-20 electrode system, with signals sampled at a rate of 500 Hz. To enhance data quality, all channels underwent bandpass filtering, with a frequency ranging from 1/1.6 Hz to 70 Hz. Certain channels, including Cz and Pz, were excluded from some recordings due to artifact constraints.

Here, we focused on the EEG data from an 8-year-old male patient. This patient experienced five recorded complex partial seizures (CPS) in the vicinity of electrode F8. The EEG data for this patient includes both an interictal segment during which no seizures are recorded and 5 ictal segments representing seizures. ,

To prepare the data for compatibility with the input structure of \sib, we divided the epileptic seizure data into non-overlapping batches, with a maximum of 8 batches extracted from each seizure. Each batch had a duration of 2000 time points, equivalent to 4 seconds. This process resulted in 4 seizures with 8 batches each and one seizure with 7 batches due to its shorter duration.

For each seizure, we also included data from the 8 seconds preceding the marked clinical identification of the seizure. This amounted to 2 additional 2000-long batches (each corresponding to 4 seconds) before each seizure event.

Regarding normal activity data, we randomly selected 40 batches, each spanning 4 seconds (2000 time points), from various time intervals that did not overlap with any seizure activity or the 8-second pre-seizure period.

In total, we had 40 batches of normal activity, 39 batches of seizure activity, and 10 batches of pre-seizure data. 

We ran SiBBlInGS on this data with $p = 7$ BBs. For the state-similarity graph ($\bm{P}$), we adopted a supervised approach to distinguish between seizure and non-seizure states, as detailed in the categorical case in Appendix~\ref{sec:supervised_equal_case}, where we assigned a strong similarity value constraint to same-state trials and lower similarity values to different-state trials.

We also leverage this example to underscore the significance of the parameter $\bm{\nu}$ in the model's ability to discover networks that emerge specifically under certain states as opposed to background networks. In this context, we defined here ${\bm{\nu} = [1,1,1,1,1,1,0]}$ such that the similarity levels of the 1st to 6th BBs are determined by the relevant values in $\bm{P}$, while the last BB's similarity is allowed to vary between states.

During the training of SiBBlInGS on this data, we adopted a training strategy where 8 random batches were selected in each iteration to ensure that the model was exposed to an equal number of trials from each state during each iteration and enhancing its robustness. 

\section{Additional Details about the Baselines}
\label{sec:comp_parafac}
\textbf{Initial Extraction of BBs from each method:}
To compare SiBBlInGS with other methods (as presented in Fig.~\ref{fig:trends_comparison},~\ref{fig:synth},~\ref{fig:epi}), we took the following approach. 
For PCA global, we applied PCA on the entire dataset after horizontally concatenating the time axis using the  sklearn~\citep{scikit-learn} implementation, specifying the number of Principal Components (PCs) to match the number of BBs in the ground truth data ($p = 10$). These PCs were then treated as the BBs.
In the case of PCA local, we followed a similar procedure. However, we ran PCA individually for each state.

For Sparse PCA global (SPCA global) and Sparse PCA local (SPCA local) we used its~\href{https://scikit-learn.org/stable/modules/generated/sklearn.decomposition.SparsePCA.html}{ sklearn implementation}, while tuning the sparsity level on the BBs (the $\alpha$ parameter) to match the sparsity level of the ground truth data.  
Similar to the regular PCA, SPCA global refers to applying a single SPCA on the entire dataset with $p=10$ components, while and Sparse PCA local (SPCA local) refers to applying SPCA to the observations of each state. 

For dPCA, we used the Python implementation offered~\href{https://github.com/machenslab/dPCA}{here}~\cite{kobak2016demixed} with $k=3$ states and $p=10$ components. We chose to protect the time axes within each trial against shuffling (``dpca.protect = ['t']'') and extracted the temporal traces using the stimulus component (``s'' key) of the trained model. 

For Tucker and PARAFAC, we utilized the Tensorly library~\citep{tensorly} with a rank set to $p = 10$ (the number of BBs allowed by SiBBlInGS). We interpreted the BBs as the first factor (factors[0] in Tensorly output), and we considered the temporal traces as the second factor (factors[1] in Tensorly output) while multiplying them by the corresponding weights from the state factor (third factor, factors[2]) to enable cross-state flexibility to these temporal traces.

For mTDR~\cite{aoi2018model, aoi2020prefrontal}, we first note that this method focuses on a slightly different problem than SiBBlInGS, specifically tailored for cases where multiple conditions influence each trial simultaneously.
Hence, in our comparative analysis, we first adapted mTDR to be comparable with SiBBlInGS by applying the following processing steps. Given that the synthetic example (Fig.\ref{fig:synth}) involves categorical rather than ordinary sequential states, we changed the categorical states to dummy variables using one-hot encoding before running mTDR. We then ran mTDR on the concatenation of all trials to obtain the temporal basis matrices ($\bm{S}$ as denoted in~\cite{aoi2018model}) and their neuron-specific weights $\bm{W}$. We recalculated the optimal coefficients based on the identified $\bm{S}$ and $\bm{W}$ to minimize the Mean Squared Error (MSE) in reconstructing each trial, and then obtained the state-specific temporal activity through optimal re-weighting of $\bm{S}$. We used the reweighed optimal $\bm{S}$ to compare it with $\bm{\Phi}$ from our paper's notation, while mTDR's $\bm{W}$ served as the structural matrices for comparison with $\bm{A}$ in our notation.
 
For NONFAT~\cite{wang2022nonparametric}, we used the code shared by the authors at~\cite{nonfat_code}. The model was executed with the same parameters as specified in~\cite{wang2022nonparametric}, but with rank set to 10 to align with the desired BBs. The algorithm was trained for 500 epochs across 10 folds. BBs were extracted from the two views of the ''$Z_{arr}$" matrix during the last epoch. The first view was reweighted using the weights obtained from the second view of ''$Z_{arr}$'' for each state and BB. Temporal traces were then extracted from the ''$U_{arr}$" matrix to calculate the trace of each BB under each state. 

For NNDTN (discrete-time NN decomposition with nonlinear dynamics, as implemented by~\cite{nonfat_code}), we concatenated individual components of ``$\bm{v}_{i_n}$'' over states over the number of BBs across all time points with re-weighting by
 ``$U_{vec}$''. The traces were then obtained by optimizing the BBs' activity to minimize the distance between the reconstruction and the original tensor.

\textbf{Post-processing steps applied to baselines' BBs and traces to align them with the ground truth:}
\begin{itemize}
    \item \textit{Synthetic Data}:

To assess and compare the results of these alternative methods against the ground truth BBs and traces, we initially normalized the BBs to fit the range of the ground truth BBs, applied sparsity using hard-thresholding such that the identified BBs from each method will present similar sparsity level to that of the ground truth, and then reordered the BBs to maximize the correlations of their temporal traces with the ground truth traces. This alignment was necessary since SiBBlInGS is insensitive to the ordering of BBs. 
For the correlation comparisons ($\rho(\bm{A}, \widehat{\bm{A}})$), we examined the correlation between the BBs, as well as their temporal traces, in comparison to the ground truth. Recognizing that correlation might not be the most suitable metric for sparse BBs comparison, we further evaluated the performance using the Jaccard index as well. 

\item \textit{EEG and Trends Experiments:}

Similar to the synthetic data scenario, in the EEG and Trends experiments, we compared the identified components with outcomes generated by different tensor and matrix factorization methods. However, since these experiments are built on real-world data, in these case, no ground truth exists for the ensembles, as the 'real' ensembles are unknown and hence the evaluation of the identified structures is qualitative. 
Specifically, after applying the baselines to the EEG/Trends data, we extracted the BBs ($\bm{A}$) by the following: In the cases of global and local PCA, these BBs were treated as the Principal Components (PCs). In the PARAFAC and Tucker tensor-factorization methods, they were considered the first factor (factors[0] from the tensorly output), weighted by the relevant components from the third factor (the states axis, factors[2]).
We then performed the following steps: 1) Normalized the matrices to ensure that each BB had a similar absolute sum of its columns, resulting in BBs of comparable magnitudes for state comparison, and 2) Introduced artificial sparsity into the matrices through hard thresholding, aiming to achieve the same level of sparsity observed in SiBBlInGS for each state.
As seen in Figure~\ref{fig:epi}, in the EEG experiment, the baselines failed to detect the emergence of BBs around electrode F8, resulting in widespread non-specific clusters; As seen in Figure~\ref{fig:trends_comparison}, in the Trneds experiment, these methods produced less meaningful BBs than SiBBlInGS.

\end{itemize}

\end{document}